\documentclass[10pt,twocolumn,letterpaper]{article}

\usepackage[pagenumbers]{iccv} 

\usepackage{booktabs}
\usepackage{xspace}
\usepackage{color}
\usepackage{amsmath}
\usepackage{pifont}
\usepackage{enumitem}
\usepackage{colortbl}
\usepackage{multirow}

\definecolor{purple}{rgb}{0.65,0,0.65}
\definecolor{blueish}{rgb}{0.0, 0.3, .6}
\definecolor{LightCyan}{rgb}{0.88,0.95,1}
\colorlet{LightCyan70}{LightCyan!70} 

\newcommand\blfootnote[1]{%
  \begingroup
  \renewcommand\thefootnote{}\footnote{#1}%
  \addtocounter{footnote}{-1}%
  \endgroup
}

\makeatletter
\renewcommand{\paragraph}{%
  \@startsection{paragraph}{4}{0pt}{2.65pt \@plus1ex \@minus.2ex}{-1em}{\normalfont\normalsize\bfseries}%
}
\makeatother

\makeatletter
\DeclareRobustCommand\onedot{\futurelet\@let@token\@onedot}
\def\@onedot{\ifx\@let@token.\else.\null\fi\xspace}

\def\eg{\emph{e.g}\onedot} 
\def\ie{\emph{i.e}\onedot}

\makeatother
\DeclareMathOperator*{\argmax}{argmax}

\newcommand{\method}{WIISA\xspace}
\newcommand{\methodextended}{Weak-labeling for Image Intrinsic Scale Assessment}
\newcommand{\dataset}{IISA-DB\xspace}
\newcommand{\rescalehyperparam}{\delta}

\definecolor{iccvblue}{rgb}{0.21,0.49,0.74}
\usepackage[pagebackref,breaklinks,colorlinks,allcolors=iccvblue]{hyperref}

\def\paperID{} 
\def\confName{ICCV}
\def\confYear{2025}

\title{Image Intrinsic Scale Assessment:\\Bridging the Gap Between Quality and Resolution}

\author{Vlad Hosu\textsuperscript{1,*} \and Lorenzo Agnolucci\textsuperscript{2,$\dagger$,*} \and  Daisuke Iso\textsuperscript{1} \and Dietmar Saupe\textsuperscript{3} \and
\textsuperscript{1} Sony AI - \tt\small[name.surname]@sony.com \\  \textsuperscript{2} University of Florence, Italy - \tt\small[name.surname]@unifi.it \\ \textsuperscript{3} University of Konstanz, Germany - \tt\small[name.surname]@uni-konstanz.de
}

\begin{document}
\maketitle

\begin{abstract}
Image Quality Assessment (IQA) measures and predicts perceived image quality by human observers. Although recent studies have highlighted the critical influence that variations in the scale of an image have on its perceived quality, this relationship has not been systematically quantified.
To bridge this gap, we introduce the Image Intrinsic Scale (IIS), defined as the largest scale where an image exhibits its highest perceived quality. We also present the Image Intrinsic Scale Assessment (IISA) task, which involves subjectively measuring and predicting the IIS based on human judgments. We develop a subjective annotation methodology and create the \dataset dataset, comprising 785 image-IIS pairs annotated by experts in a rigorously controlled crowdsourcing study with verified reliability. Furthermore, we propose \method (\methodextended), a strategy that leverages how the IIS of an image varies with downscaling to generate weak labels. Experiments show that applying \method during the training of several IQA methods adapted for IISA consistently improves the performance compared to using only ground-truth labels. The code, dataset, and pre-trained models are available at \small{\href{https://github.com/SonyResearch/IISA}{\url{https://github.com/SonyResearch/IISA}}}.
\end{abstract}

\blfootnote{\textsuperscript{$*$}~Equal contribution.} 
\blfootnote{\textsuperscript{$\dagger$}~Work done during an internship at Sony AI.} 

\begin{figure}[!t]
    \centering
    \includegraphics[width=\columnwidth]{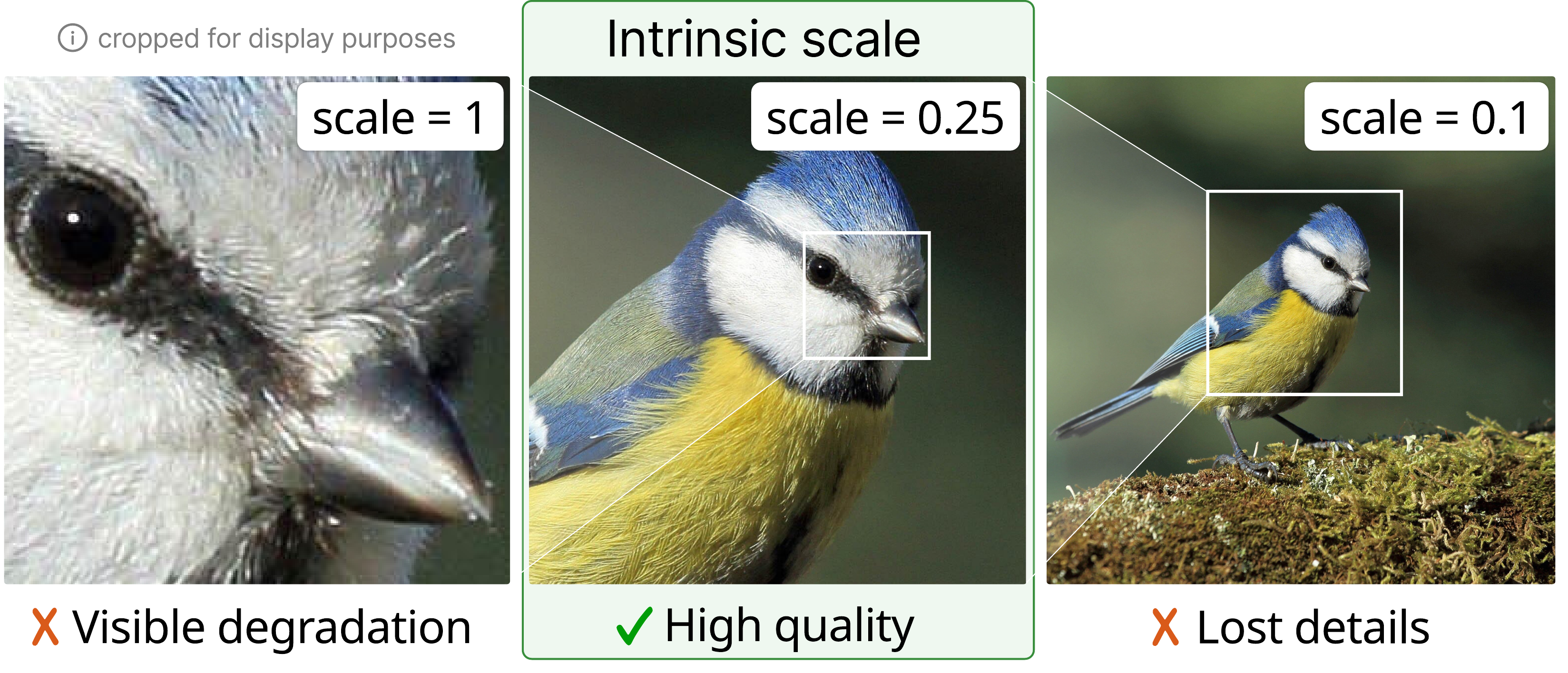} \\
    \vspace{10pt}
    \includegraphics[width=0.7\columnwidth]{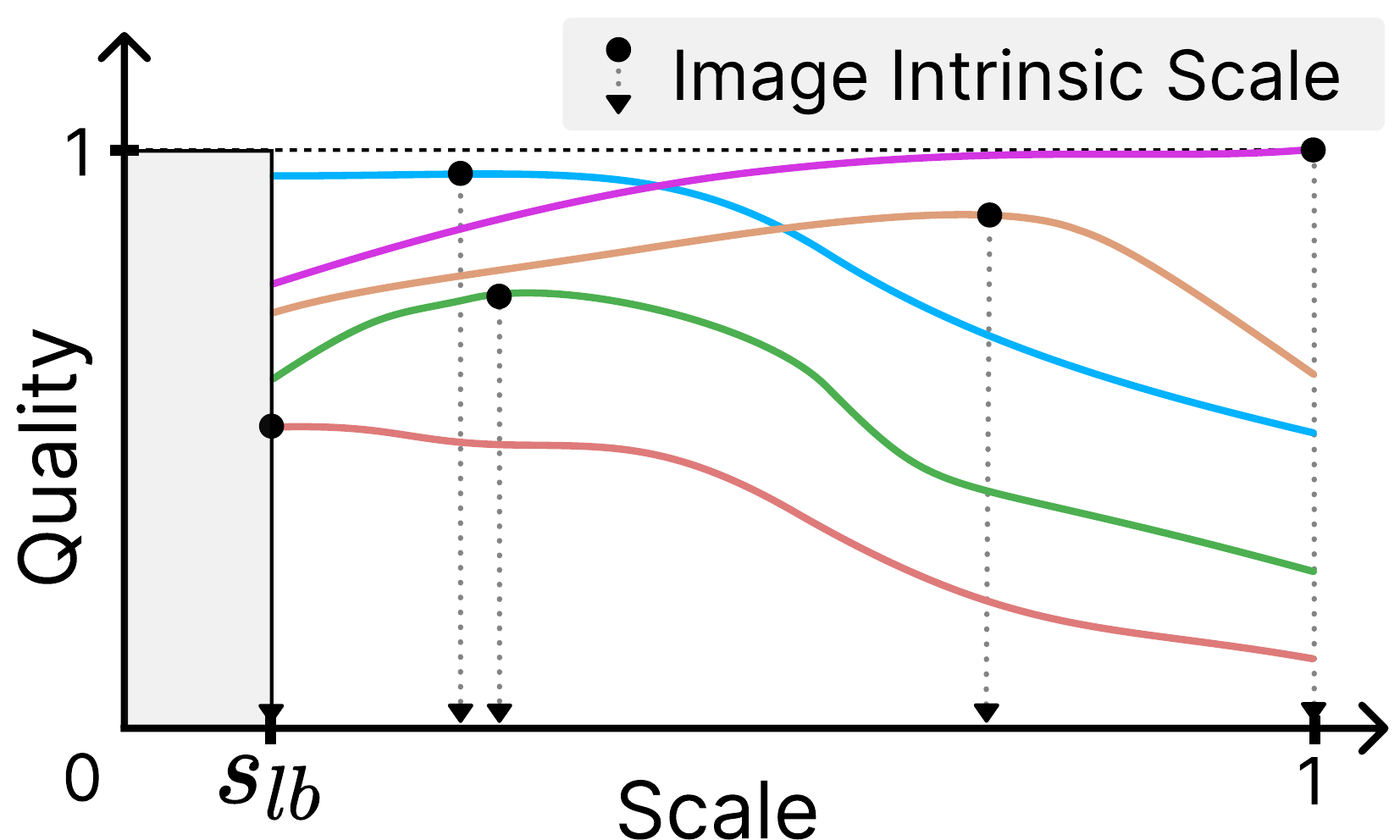}
    \caption{
    \textbf{Top:} downscaling an image affects its perceived quality. As the original image (left) is downscaled, degradation becomes less noticeable, but some high-frequency details may be lost. The optimal trade-off occurs at the image's intrinsic scale. Note that the first two images are cropped for clearer visualization.
    \textbf{Bottom:} we plot several possible profiles corresponding to different images, depicting how the quality changes with scale -- quality refers to the latent expected value of measured subjective quality ratings rather than the noisy mean observed from finite samples. The Image Intrinsic Scale (IIS) is the largest scale at which an image shows its highest perceived quality. The IIS is measured above a lower bound $s_{lb}\!=\!0.05$, where the image quality is more interpretable.
    }
    \label{fig:teaser}
    \vspace{-10pt}
\end{figure}

\section{Introduction} \label{sec:intro}
Image Quality Assessment (IQA) aims to model and quantify subjectively how the degradation of an image, such as that introduced by poor camera performance or image processing, affects its perceived quality. Particularly, Full-Reference IQA (FR-IQA) evaluates quality by comparing a pristine reference image with its degraded counterpart, while No-Reference IQA (NR-IQA) assesses quality without needing a reference image. Despite significant advancements in IQA, the interaction between image quality and spatial resolution remains a critical under-explored area. While the resolution, represented by Megapixels, is often seen as a key indicator of image quality in cameras, it does not account for the degradations introduced by the camera system, such as lens diffraction or increased noise due to smaller pixel size in higher resolution sensors. Paradoxically, higher resolution acquisition can lead to lower perceived quality. Thus, further research is needed to explore how resolution impacts image quality perception.

Most IQA datasets annotate quality ratings -- in the form of Mean Opinion Scores (MOS) -- for images at a fixed resolution \cite{hosu2020koniq, hosu2024uhd, fang2020perceptual}. To the best of our knowledge, KonX \cite{wiedemann2023konx} was the first dataset that provided quality annotations for each image rescaled across multiple resolutions. The authors showed that perceived quality varies with image resolution (\ie, size in pixels).
When an image is downscaled, its resolution decreases. Since images are typically displayed at a consistent pixel density on any given display medium, downscaled images also appear physically smaller. The perceived quality of smaller images often increases because degradation becomes less noticeable, e.g., noise grain becomes smaller or fades. Conversely, relevant high-frequency details may get lost when the resolution is too low, leading to reduced perceived quality. The former phenomenon is exemplified in \cref{fig:teaser} (top): an image that appears blurry at its original size ($scale=1$) becomes crisp and clear when downscaled to $scale=0.25$, such as the bird's feathers. However, at a smaller scale ($scale=0.1$), fine details such as individual barbs are lost.

Based on the earlier observation, we aim to answer the question: 
\textit{what is the optimal scale of an image that provides the best balance between visibility of degradation and retention of relevant high-frequency details?} Thus, we introduce the concept of \textbf{Image Intrinsic Scale} \textbf{(IIS)}, which is the largest scale at which an image shows its highest perceived quality. We select the largest scale as it maximizes the availability of high-quality information. Moreover, we propose a new task named Image Intrinsic Scale Assessment (IISA). IISA aims to subjectively measure and predict the IIS in alignment with human perception. Our main goal is to develop effective automated methods for predicting the IIS, and this first requires establishing a reliable annotation methodology to collect subjective opinions.

Subjective studies measure latent variables that are not directly observable -- here, the perceived quality or IIS. These latent variables are commonly modeled as random variables, with each human opinion representing a sample. The theoretical “true” value is the expectation of this random variable. In principle, when we refer to image quality or IIS, we discuss the expected value, which can be estimated by averaging many independent and unbiased opinions, \eg, via the MOS. In practice, the number of opinions is limited by cost and other constraints, so we work with approximations that depend on careful experimental design to minimize bias and achieve reliable estimates.

To collect ground-truth IIS labels, we propose an annotation methodology that leverages a tailored user interface that we developed. Our annotation tool allows the participants to intuitively downscale the original image via a slider until it reaches its IIS. We ask every annotator to label each image twice, a few days apart, ensuring excellent self-consistency and reliability. The final ground-truth IIS labels are obtained by aggregating each participant's individual subjective opinions, similar to the process of calculating the MOS for IQA. Based on the proposed annotation methodology, we create the \dataset dataset, which contains 785 image-IIS pairs labeled via crowdsourcing. We select freelance image quality experts as participants to enhance the reliability of ground-truth IIS labels.

To automatically predict the IIS, we initially assess the zero-shot transfer capabilities of pre-trained NR-IQA models. The subpar performance of these models on IISA underscores the distinct nature of the two tasks and the need for IISA-specific training.
To this end, we propose \method (\methodextended), a straightforward yet effective approach for generating multiple weakly labeled image-IIS pairs from a single ground-truth one. \method leverages a unique property of the intrinsic scale that allows us to extrapolate the ground-truth IIS of an image to its downscaled versions via a piece-wise function. Our method can be easily integrated into the training of any model, as it is independent of specific training strategies or model characteristics. The experiments demonstrate that the proposed approach consistently boosts the performance of several methods from the NR-IQA domain retrained for IISA on the proposed dataset.

We summarize our contributions as follows:
\begin{itemize}
    \item We define the concept of Image Intrinsic Scale (IIS), which is the largest scale at which an image shows its highest perceived quality.
    \item We propose a new task, the Image Intrinsic Scale Assessment (IISA), aiming to subjectively measure and predict the IIS in alignment with human perception.
    \item We design an annotation methodology for collecting subjective opinions of the IIS that ensures high participant self-consistency and reliability of the labels. Moreover, we create the \dataset dataset, comprising 785 image-IIS pairs annotated by image quality experts in a thoroughly controlled and reliable crowdsourcing study.
    \item We propose \method to generate multiple weakly labeled image-IIS pairs from a single ground-truth pair.
\end{itemize}

\section{Related Work}

Prior research has examined concepts related to the IIS, namely camera performance metrics, image quality under upscaling, the relationship between quality and viewing distance, and no-reference image quality assessment.

\paragraph{Camera performance metrics} The intrinsic scale is fundamentally related to a physical property of camera and lens systems: their resolving power or ability to discern details in an image. This property is theoretically limited by the camera's sampling resolution, but in practice, it is often lower due to imperfections in the lenses and imaging system. Traditionally, these properties have been studied for the camera as a whole using metrics such as Perceptual Megapixels (P-MP) \cite{dxomark2012perceptual} and the closely related Modulation Transfer Function (MTF) \cite{boreman2001}, leading to the concept of ``intrinsic camera resolution'' \cite{burns2015intrinsic}. However, because previous approaches characterize cameras under ideal laboratory conditions, they do not generalize well to real-world photos, which are affected by a multitude of unaccounted degradations. Therefore, for the first time, we propose to study the resolving power conditioned on individual images by subjectively measuring the IIS. This approach allows our method to generalize effectively to real-world photographs.

\paragraph{Quality under upscaling} Another line of research has focused on the effects of upscaling on image quality.
Works such as \cite{shah2021real, zhu2021perceptual} aim to discriminate between authentic and artificially upscaled images. While valuable for detecting upscaling artifacts, such a binary classification does not offer a generalizable image quality metric.
\citet{kansy2023self} introduce the concept of ``effective resolution", \ie, the smallest size an image can be downscaled to, such that when it is upscaled back to its original dimensions the loss of detail is not perceptible.
This is distinct from the IIS. Consider the example of a noisy image: the scale of its effective resolution should be \textit{high} enough to preserve the noise pattern such that upscaling will not lead to noticeable detail loss relative to the original. In contrast, the IIS of the image should be \textit{low} enough to make the noise not visible anymore, improving the perceived quality of the image.

\paragraph{Quality and viewing distance} Several works studied the impact of viewing distance on the perceived quality of images \cite{liu2014cid, gu2015quality, fang2015evaluation} and videos \cite{hammou2024effect, keller2023influence, kufa2019visual}. \citet{fang2015evaluation} collected subjective opinions on image Quality of Experience (QoE) across seven viewing distances, showing that QoE generally increases with distance due to reduced visibility of distortions but declines when image details become less discernible at greater distances. This results in a concave-down relationship between QoE and viewing distance, aligning with findings by the authors of the KonX dataset \cite{wiedemann2023konx} on the quality-resolution relationship. Based on these insights, in \cref{sec:iisa} we establish a rule to extrapolate the IIS of an image to its downscaled versions.

\paragraph{No-Reference Image Quality Assessment}
Although the IIS differs significantly from traditional quality ratings, it is closely connected through the underlying subjective judgments. The IIS is derived from the quality assessments of all downscaled versions of an image. Consequently, No-Reference Image Quality Assessment (NR-IQA) methods that predict quality ratings from single images are related.

In recent years, NR-IQA has attracted significant attention, particularly in the development of supervised methods \cite{su2020blindly, ke2021musiq, chen2024topiq, golestaneh2022no, wiedemann2023konx}. A notable trend within these approaches is the usage of multi-scale representations \cite{ke2021musiq, wiedemann2023konx, chen2024topiq}, which have proven effective in enhancing cross-resolution generalization. For instance, TOPIQ \cite{chen2024topiq} employs an attention-based network to extract top-down multi-scale features while propagating high-level semantic information.
Another line of research leverages self-supervised learning to train an image encoder on unlabeled data, followed by regressing the encoder’s features to predict quality scores \cite{madhusudana2022image, agnolucci2024arniqa, zhao2023quality, saha2023re}. For example, CONTRIQUE \cite{madhusudana2022image} involves maximizing the similarity between the representations of crops coming from the same distorted image via a contrastive loss.
More recent works \cite{wang2023exploring, agnolucci2024quality, zhang2023blind} leverage vision-language models like CLIP \cite{radford2021learning} to measure the quality of an image based on its similarity to antonym prompts.

\section{Image Intrinsic Scale Assessment} \label{sec:iisa}

As illustrated in \cref{fig:teaser} (bottom) and supported by previous studies \cite{wiedemann2023konx, hammou2024effect, fang2015evaluation, liu2014cid, gu2015quality}, the relationship between the perceived quality of an image and its presentation scale is generally assumed to follow a concave-down or monotonic function for scales smaller than 1 (\ie, the original size). The ``presentation scale'' is the scale at which the image is displayed for subjective annotation. Downscaling an image initially improves quality by reducing degradations like blur and noise. However, excessive downscaling diminishes quality due to detail loss and artifacts like aliasing. 

To quantify the relationship between quality and scale, we introduce the IIS, \ie, the largest scale at which an image shows its highest perceived quality. The IIS stems from subjective evaluations of the quality of an image's downscaled versions. Therefore, analogous to the MOS in IQA, ground-truth IIS labels are determined by aggregating the individual subjective judgments from multiple annotators into a single scalar value. Based on the concept of IIS, we propose the IISA task, which has two main aspects:
\begin{enumerate}
    \item \textit{Subjective}: subjectively measuring the IIS (see \cref{sec:dataset});
    \item \textit{Predictive}: developing automatic methods for predicting the IIS in alignment with human judgments (see \cref{sec:approach}).
\end{enumerate}
Although our annotation study presents a single image at a time, the IIS is determined by comparing the quality of rescaled versions of an image. Thus, the subjective side of IISA has a mixed nature, including comparisons across scales. In contrast, the predictive task is strictly no-reference, using a single image to predict the IIS.

Formally, let $I$ be an image. Denote $I^s$ as the image obtained by downscaling $I$ to a scale $s$, with $0 < s \leq 1$. Let $Q(I^s)$ be the quality of $I^s$, measurable by the MOS. Then, we define the IIS of $I$ as $\Omega(I)$, which gives the maximum scale at which $I$ shows its highest perceived quality:
\begin{equation} \label{eq:iis_definition}
\begin{gathered}
    \Omega(I) = \max( \argmax_{s_{lb} \leq s \leq 1}( Q(I^s) ) ) .
\end{gathered}
\end{equation}
In our implementation, we set a lower bound $s_{lb}=0.05$ for the scale $s$ because reliably evaluating the quality of very small images is challenging due to the inherent loss of detail from downscaling. 
When assessing image quality, slight variations in size typically result in minimal quality change. Therefore, an image might be perceived as having its highest quality at multiple scales. Thus, we take the maximum value among these scales in \cref{eq:iis_definition}, as it maximizes the presence of high-quality information.

When downscaling images, different interpolation methods introduce artifacts like aliasing, blurring, or ringing, which affect the IIS. To achieve the highest quality, we use Lanczos interpolation. Its ability to preserve fine details and minimize aliasing via sinc-based filtering results in higher image quality compared to bilinear or bicubic methods. Consequently, we employ this method for both the subjective and predictive aspects of IISA.

\begin{figure}[t]
    \centering
    \includegraphics[width=0.7\columnwidth]{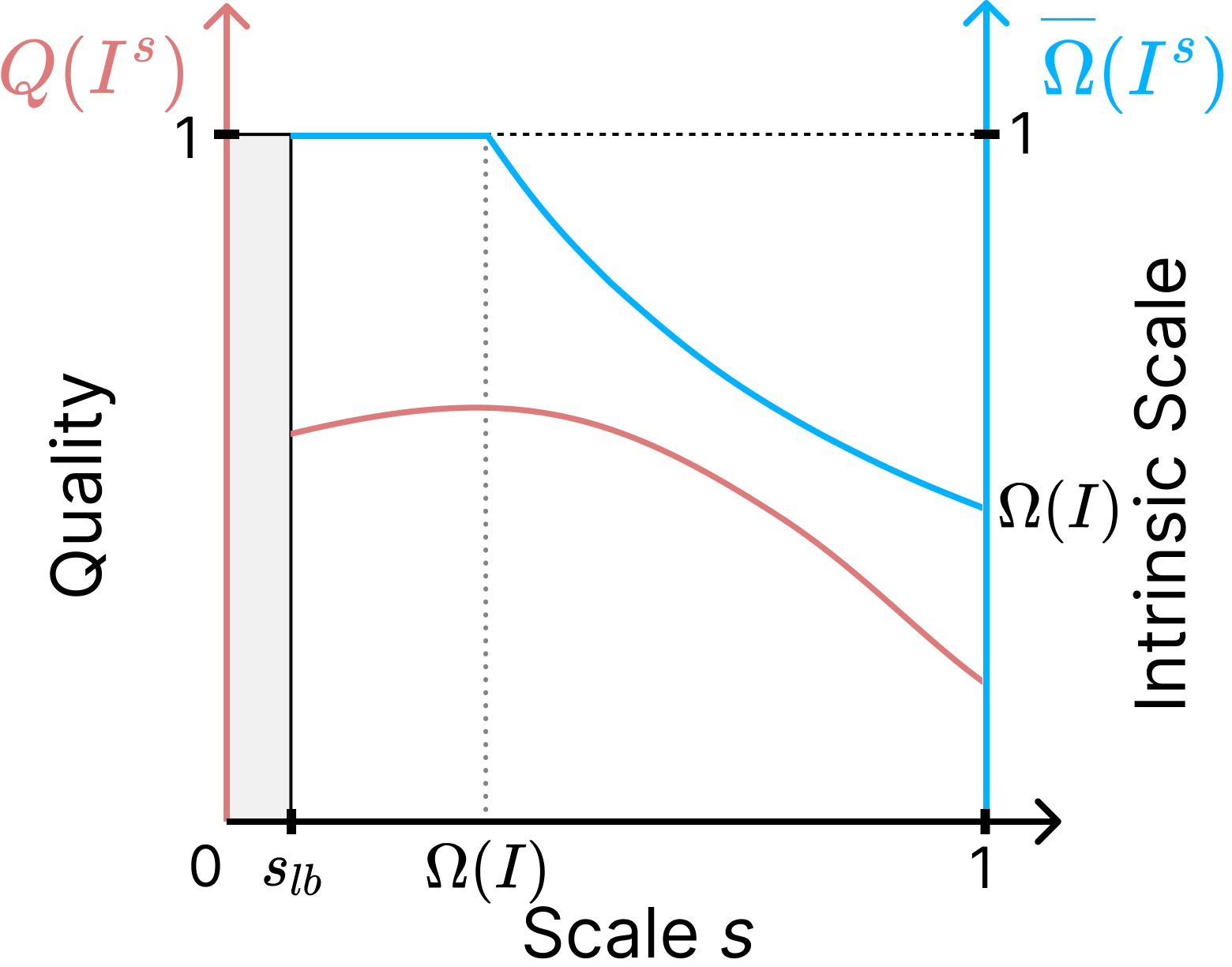} 
    \vspace{-5pt}
    \caption{
    Example of the influence of the scaling factor $s$ on the quality $Q(I^s)$ of downscaled images $I^s$ and the intrinsic scale $\overline{\Omega}(I^s)$, based on \cref{eq:iis_formula}. The quality is not evaluated below the lower bound $s_{lb}$. $\Omega(I)$ is the scale where the quality function is at its maximum, \ie, the IIS of the original image $I$ ($s=1$).
    }
    \vspace{-10pt}
    \label{fig:iis_formula}
\end{figure}

Given the ground-truth IIS of an image, we can easily extrapolate it to its downscaled versions by applying the definition in \cref{eq:iis_definition}. To do so, we start from the assumption that the quality of downscaled images $Q(I^s)$ is a concave-down or monotonic function of scale, thus:
\begin{enumerate}[label=\textit{A\arabic*:}, left=1em]
    \item $Q(I^s)$ increases monotonically for $s \leq \Omega(I)$;
    \item $Q(I^s)$ decreases monotonically for $s > \Omega(I)$. 
\end{enumerate}
Then, starting from the ground-truth IIS $\Omega(I)$ of an image $I$, we can infer the IIS $\overline{\Omega}(I^s)$ of the downscaled versions of $I$ via a piece-wise function:
\begin{equation} \label{eq:iis_formula}
    \overline{\Omega}(I^{s}) = \begin{cases} 
      1 & s_{lb} \leq s \leq \Omega(I) \\
      \frac{\Omega(I)}{s} & \Omega(I) < s \\
   \end{cases} .
\end{equation} 
From now on, we use $\Omega$ to refer to the ground-truth IIS, while with $\overline{\Omega}$ we indicate intrinsic scales obtained via \cref{eq:iis_formula}. We can consider the IIS as the minimum amount we need to downscale an image to maximize its quality. Therefore, starting from a downscaled image $I^s$ and seeking its IIS, \cref{eq:iis_formula} can be interpreted as follows:
\begin{itemize}[leftmargin=*]
    \item First branch: when the initial scale $s$ is below or at $\Omega(I)$, due to the assumption \textit{A1}, downscaling $I^s$ further cannot increase its quality, as it is already at its maximum. Therefore, by the definition of the intrinsic scale (\cref{eq:iis_definition}), the IIS of $I^s$ is its original scale, \ie, 1, so $\overline{\Omega}(I^s)=1$;
    \item Second branch: when the initial scale $s$ of $I^s$ is above $\Omega(I)$, due to the assumption \textit{A2}, the quality is at its highest at the scale $\Omega(I)$. Hence, the amount of downscaling required to bring $I^s$ to its IIS is given by the ratio of the original intrinsic scale and the scale $s$, \ie, $\overline{\Omega}(I^s)=\frac{\Omega(I)}{s}$.
\end{itemize}
\cref{fig:iis_formula} shows an example of the application of \cref{eq:iis_formula} to the downscaled versions $I^s$ of a generic image $I$.

Therefore, based on \cref{eq:iis_formula}, we can infer the IIS of the downscaled versions of a labeled image. This idea underlies the proposed approach, named \method, which generates multiple weak labels from a single image-IIS pair. The experimental results, reported in \cref{sec:experimental_results}, show that this strategy significantly improves predictive performance.

\paragraph{Differences with NR-IQA}
IISA differs from NR-IQA in several key aspects. First, IISA offers a concrete way to maximize an image's perceived quality by simply downscaling it to its intrinsic scale. In contrast, NR-IQA only provides information on an image's quality at a fixed scale.
Second, compared to NR-IQA, IISA is affected less by global image degradation, such as overall contrast and color defects. The downscaling operation mainly affects spatially limited defects such as blurs, compression artifacts, and noise. 
Third, NR-IQA lacks a definitive model for how image quality changes with scale. While studies like \cite{hammou2024effect} explore the relationship between perceived quality and viewing distance, accurately understanding this change requires annotating an image (\eg, MOS) at multiple scales. The KonX dataset \cite{wiedemann2023konx} provides ratings for three scales, but fine-grained sampling across all scales is impractical. On the contrary, for IISA, we can easily extrapolate the ground-truth IIS of an image to its downscaled versions by using \cref{eq:iis_formula}. 
Fourth, quality ratings use a perceptually linear scale \cite{devellis2021scale}, whereas rescaling factors underlying the IIS are inherently non-linear. For example, despite the same absolute difference, changing the scale from 0.2 to 0.1 halves the image size, whereas decreasing it from 0.6 to 0.5 results in only a 17\% drop.
Fifth, the quality MOS in NR-IQA struggle with limited sensitivity. This is particularly true when evaluating subtle quality differences in high-quality images resulting from restoration, enhancement, or compression methods. This has led the JPEG standards committee to explore techniques capable of more granular quality assessment \cite{testolina2023jpeg}. 
In the supplementary material, we prove that IISA significantly outperforms traditional NR-IQA quality ratings in detecting subtle differences, exhibiting several times greater sensitivity.

\paragraph{Applications} 
IISA addresses applications where understanding scale-dependent quality is critical:
1) it optimizes image display and storage by identifying an image's peak quality across scales, crucial for applications like the web, print, and gaming. This enhances user experience and saves storage or printing space by using the display media's pixel density (\eg, PPI) to determine optimal display sizes.
2) IISA improves quality measurement sensitivity over traditional NR-IQA methods because it relies on implicit cross-scale comparative judgments. This makes it effective for detecting subtle quality degradations from restoration methods like denoising and super-resolution. See the supplementary material for more details on sensitivity.
3) IISA supports the creation of better low-level vision datasets. Unlike current methods that reduce diversity by discarding low-quality images, by downscaling high-resolution images to their IIS, IISA can help maintain both quality and diversity, essential for diverse and high-quality training datasets \cite{gu2019div8k, li2023lsdir}. See the supplementary material for more details on the applications.

\section{\dataset Dataset Creation} \label{sec:dataset}

We address the subjective task of IISA, defining an annotation methodology to measure the IIS subjectively and collect ground-truth annotations, which results in our \dataset dataset. Refer to the supplementary material for more details about the dataset.

\subsection{Dataset collection}
The subjective side of IISA resembles the image Just Noticeable Difference (JND) task \cite{jin2016statistical, shen2020jnd, lin2022large}, which identifies the minimal distortion level that becomes noticeable with increased image compression. Similarly, in IISA, the goal is to find the lowest downscaling factor that maximizes the perceived quality. The analogy with JND inspired the design of our annotation tool.

\paragraph{Annotation tool} 
We developed Zoom Viewer (ZOVI), a web application for annotating the IIS of images and facilitating web-based crowdsourcing experiments through batch image presentation. ZOVI features a slider for users to adjust the image scale from $s=1$ to $s_{lb}=0.05$ to identify the IIS. This slider method, inspired by JND research \cite{lin2022large}, reduces annotation effort and enhances reliability. The images are rescaled using the standard Lanczos interpolation and can be panned if the display area is insufficient. ZOVI displays each image starting at its original resolution ($scale = 1$), each image pixel being displayed at the same size as native screen pixels (1-to-1 ratio). Participants were instructed to downscale each image until no quality improvement was observed and to stop at the largest scale where this held true, thereby determining the individual IIS.

\paragraph{Annotation process}
We collected 20 opinions per image from 10 freelance participants, meeting the minimum of 15 per stimulus recommended by \cite{RecommendationITUT913} for quality assessment. Given the satisfactory confidence intervals reported in \cref{sec:dataset_analysis}, the number of participants is adequate. This aligns with similar IQA studies like SPAQ \cite{fang2020perceptual} and UHD-IQA \cite{hosu2024uhd}, which gather 15 and 20 opinions per image, respectively. All freelancers had previous experience annotating IQA datasets and a professional background in visual arts, such as photography and graphic design. Before starting the main task, participants had to undergo a training phase. Specifically, they annotated images with predefined ground truth IIS ranges determined by the authors of this paper, with textual hints provided to guide accurate annotation. After the training stage, participants annotated images presented in randomly ordered batches of 90. Each batch was viewed twice, with a few days between repetitions. We measured Spearman's Rank-order Correlation Coefficient (SRCC) between the annotations of repetitions of corresponding batches for each participant. Pairs of batches with an SRCC lower than 0.5 were required to be re-annotated. This process ensures accurate individual annotations and an overall high experiment reliability.

\paragraph{Image selection}
Following our annotation strategy, we created the \dataset dataset, comprising 785 image-IIS pairs labeled at their original resolution via crowdsourcing. All images are larger than 2048 pixels in width. We started from an initial set of 900 authentically distorted real-world images. We sampled 300 of them from the KonIQ-10k \cite{hosu2020koniq} IQA dataset according to their MOS and machine tags to ensure a diverse range of quality levels and content types. The remaining 600 images were sourced from the \textit{Pixabay} website and selected for diversity based on multiple factors, such as user tags and number of likes. 
Finally, to address privacy concerns, we manually removed any images containing identifiable individuals, resulting in the final set of 785 images. 
As a result of the sampling procedure, our dataset contains images with diverse subjects, quality levels, resolutions, and consequently, IIS values.

\paragraph{Ground-truth labels}
We obtained the ground truth IIS labels for each image (\ie $\Omega(I)$) by aggregating individual annotations from all participants, similar to how Mean Opinion Scores (MOS) are calculated for image quality ratings. Therefore, we refer to these aggregated labels as the Mean Opinion Intrinsic Scale (MOIS). This aggregation effectively reduces the variability in perceived quality due to the diverse viewing conditions inherent in crowdsourcing. Considering the non-linear relationship between scale values in our slider UI -- where different lengths on the slider represent equal ratios between two scale values -- we chose the geometric mean to compute the MOIS. 

\subsection{Dataset Analysis} \label{sec:dataset_analysis}

\paragraph{Reliability of the annotations}
To assess the reliability of the subjective annotations of \dataset, we compute the confidence intervals (CIs) for the MOIS of each image. Following a well-established methodology, we sample 20 opinions with replacement for each image, compute their geometric mean, and repeat the process 100 times. The 95\% CI is then estimated as half the range between the $2.5^{\text{th}}$ and $97.5^{\text{th}}$ percentiles of the simulated geometric means. This yields an average CI of $0.057$ for our dataset. As a reference, we consider the KonX NR-IQA dataset, whose annotation conditions were similar to ours, involving remote studies with freelance participants. The average CI of the MOS for the KonX dataset is $0.046$. Given the differences between the IISA and NR-IQA tasks, some variation in the CIs is expected. Therefore, this outcome indicates that our annotation process has a level of reliability comparable to a highly reliable NR-IQA dataset such as KonX.

\begin{figure}[t]
    \centering
    \includegraphics[width=\columnwidth]{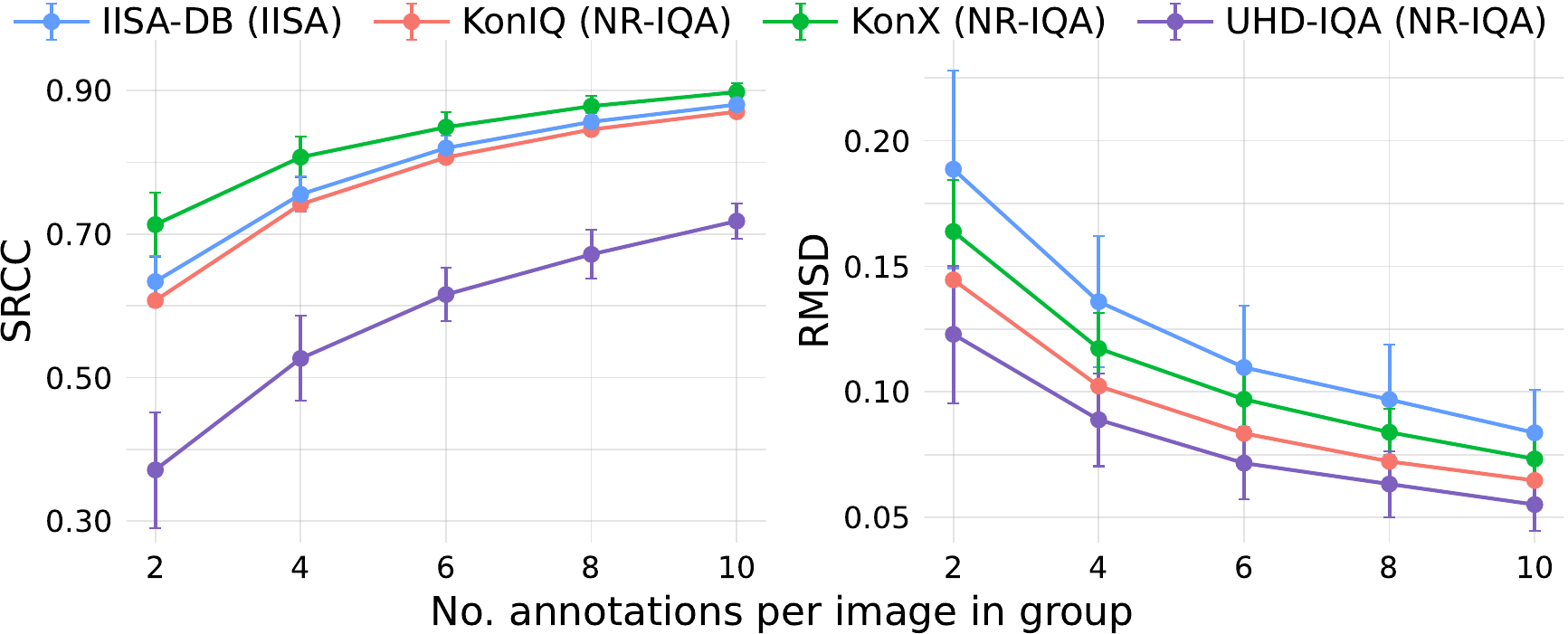}
    \vspace{-15pt}
    \caption{Comparison of inter-group agreement between the proposed \dataset and existing NR-IQA datasets, across different numbers of annotations per image in each group. The dots represent the average values, while the error bars indicate the range of $\pm$1 standard deviation.
    }
    \vspace{-10pt}
    \label{fig:intergroup_agreement}
\end{figure}

We further assess the reliability of our annotations using a common approach that measures the inter-group agreement between different participant sub-groups and compares it with that of existing NR-IQA datasets \cite{hosu2024uhd, wiedemann2023konx, hosu2020koniq}. Specifically, for datasets where each participant annotated each image twice (\dataset, KonX \cite{wiedemann2023konx}, UHD-IQA \cite{hosu2024uhd}), we randomly sample without replacement pairs of non-overlapping groups of up to five participants each, thus considering at most ten annotations per image. For KonIQ-10k \cite{hosu2020koniq}, since participants provided just a single opinion per image and each rated only a subset of images, we sample the required number of opinions per group without considering participant identity. For each dataset, we sample 200 pairs of groups and compute the average SRCC and RMSD (Root Mean Squared Difference) between the MOIS (IISA) or MOS (NR-IQA) of each group. Note that although both IIS and quality ratings are scaled to the $[0,1]$ range, their absolute errors are not directly comparable due to the different nature of their scales -- non-linear and linear, respectively. Therefore, SRCC is more appropriate than RMSD in comparing inter-group agreements. Nonetheless, we report both for completeness in \cref{fig:intergroup_agreement}. We observe that the proposed \dataset dataset has SRCC values that are comparable to existing reliable NR-IQA datasets, such as KonIQ-10k. The same applies to the RMSD. Overall, these results demonstrate the reliability of our dataset's annotations.

\paragraph{Effects of variable viewing conditions}
Viewing conditions -- including display characteristics (contrast, brightness), environmental lighting, viewing distance, and viewer acuity -- significantly influence perceived image quality. Laboratory experiments tightly control these factors but consequently measure atypical conditions with limited variability, restricting the applicability of the findings. In contrast, crowdsourcing involves diverse and less controlled conditions, potentially reducing individual participant reliability. Nevertheless, when reliable group opinions are obtained through well-controlled methods, the resulting assessments generalize better. Our annotation framework has demonstrated high reliability, ensuring that, despite individual variation, group aggregates accurately measure the perceptual IIS. Thus, we provide highly reliable data that better reflects real-world viewing conditions, enhancing the validity and applicability of our findings.

\begin{table}
    \centering
    \Large
    \resizebox{\linewidth}{!}{ 
    \begin{tabular}{lcccc}
    \toprule
    Method & SRCC & PLCC & RMSE & MAE \\ \midrule
    TOPIQ (SPAQ) & 0.475 & 0.437 & \textbf{0.136} & \textbf{0.104} \\
    TOPIQ (UHD-IQA) & \textbf{0.587} & \textbf{0.600} & 0.382 & 0.364 \\ \bottomrule
    \end{tabular}}
    \vspace{-5pt}
    \caption{Evaluation of the zero-shot transfer performance of pre-trained NR-IQA models on the \dataset dataset.  ($\cdot$) indicates the pre-training dataset. Best scores are highlighted in bold.}
    \label{tab:naive_results}
    \vspace{-10pt}
\end{table}

\section{IISA Predictive Models} \label{sec:approach}

In this section, we focus on the prediction task of IISA, \ie, developing methods to automatically predict the IIS in alignment with human perception. First, we evaluate the zero-shot transfer performance of pre-trained NR-IQA models on our dataset. We then introduce \method, an approach for generating weak IIS labels that can be applied during the training of any IISA model.

\subsection{Pre-trained NR-IQA Models} 
The IIS is derived from quality ratings across all downscaled versions of an image. It primarily depends on low-level degradations like blur and noise, not aesthetics or semantics. These reasons make IISA and NR-IQA closely related, despite their several key differences (see \cref{sec:iisa}). Moreover, since NR-IQA methods evaluate both distortions and high-quality details, they are plausible candidates for the IISA task. Therefore, we evaluate the performance of pre-trained NR-IQA models for zero-shot IIS prediction.

We consider TOPIQ \cite{chen2024topiq}, a state-of-the-art NR-IQA approach with strong generalization capabilities. We pre-train it on the SPAQ \cite{fang2020perceptual} and UHD-IQA \cite{hosu2024uhd} NR-IQA datasets, which feature high-resolution images, similar to \dataset. We avoid using KonIQ-10k \cite{hosu2020koniq} due to image overlaps with our dataset. We apply the pre-trained TOPIQ models to predict quality scores on \dataset, which we directly use as IIS estimates as both metrics range in $[0,1]$. We follow the evaluation protocol described in \cref{sec:evaluation_protocol} and present the results in \cref{tab:naive_results}. Both models achieve unsatisfactory performance, highlighting the inefficacy of pre-trained NR-IQA models for zero-shot IIS prediction. This emphasizes the need for IISA-specific training and techniques.

\subsection{Proposed Approach: \method}

\begin{figure}
    \centering
    \includegraphics[width=\columnwidth]{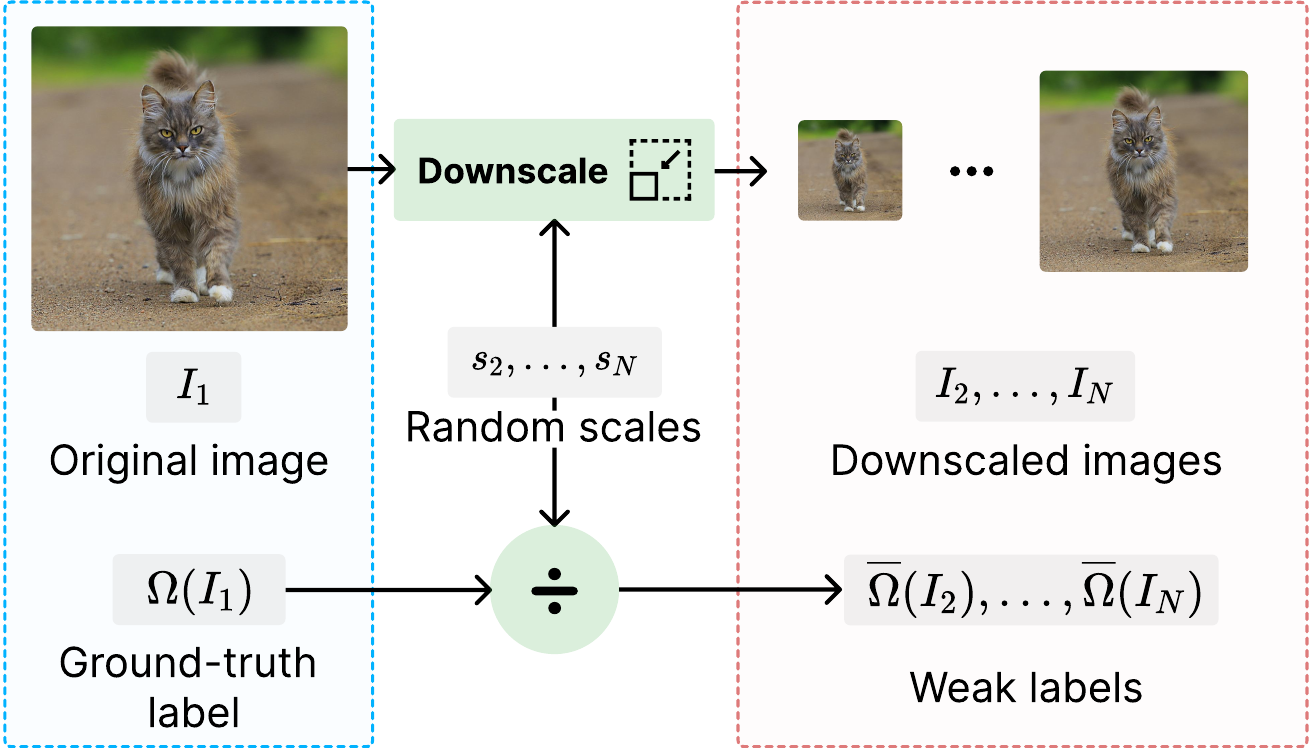} 
    \caption{Overview of \method, the proposed weak-label generation strategy. Given a ground-truth image-IIS pair ($I_1$, $\Omega(I_1)$) (blue dotted line), we randomly sample multiple scale values. Then, we divide the ground-truth label $\Omega(I_1)$ by the random scales (following \cref{eq:iis_formula}) to get the weak IIS labels associated with the downscaled versions of the original image $I_1$ (red dotted line).}
    \label{fig:pseudo_labels_generation}
\end{figure}

Given an image and its labeled IIS, we can easily leverage \cref{eq:iis_formula} to extrapolate the IIS to rescaled image versions. Based on this idea, we propose a weak-label generation approach named \method (\methodextended). In particular, starting from each labeled image-IIS pair, we randomly downscale the image to multiple scales. Then, we use \cref{eq:iis_formula} to compute the IIS associated with each downscaled version of the original image to obtain a weakly labeled image-IIS pair. Our approach significantly enhances the value of each annotation by generating additional training data from it. Moreover, since it depends solely on the properties of the IIS, \method can be employed to enhance the training of any IISA model, regardless of its specific characteristics and training strategy.

Formally, let $I_1$ and $\Omega(I_1)$ be an image and its corresponding ground-truth IIS, respectively. We randomly sample $n_{wl}$ scales in the range $[\max(\Omega(I_1), \rescalehyperparam), 1]$, where $\rescalehyperparam$ is a threshold hyperparameter empirically set to 0.65. In our experiments, we use ${n}_{wl}=2$ unless stated otherwise. Since the scales are larger than $\Omega(I_1)$, they fall in the second branch of \cref{eq:iis_formula}. Then, we downscale $I_1$ based on each of these scales using Lanczos interpolation. Therefore, given $N = 1 + n_{wl}$ and the set of randomly sampled scales $\mathcal{S}=\{s_i \mid i=2,\dots,N \} $, we get a corresponding set of downscaled versions of the same image $\mathcal{I}=\{I_i \mid i=2,\dots,N \}$. Finally, we leverage \cref{eq:iis_formula} to obtain the corresponding set of IIS $\overline{\mathcal{S}_{\Omega}}=\{\overline{\Omega}(I_i) \mid i=2,\dots,N \}$. Thus, given $i=2,\dots,N$, $I_i\in\mathcal{I}$ represents the image obtained by rescaling $I_1$ to a scale $s_i \in \mathcal{S}$ (\ie, $I_{1}^{s_{i}}$), with an associated IIS of $\overline{\Omega}(I_i) \in \overline{\mathcal{S}_{\Omega}}$. In the end, starting from a single ground truth image-IIS pair ($I_1$, $\Omega(I_1)$), thanks to our approach, we get $n_{wl}$ weakly labeled pairs ($I_i$, $\overline{\Omega}(I_i)$), with $i=2,\dots,N$. \cref{fig:pseudo_labels_generation} shows an overview of the proposed strategy. During training, we use \method to augment each batch of size $B$ with additional $B \cdot n_{wl}$ weakly labeled training samples. Since our approach only requires image downscaling to generate each weak label, it can be easily applied online during training.

\section{Experimental Results} \label{sec:experimental_results}

\subsection{Evaluation Protocol} \label{sec:evaluation_protocol}
We measure the performance using Spearman's Rank-order Correlation Coefficient (SRCC), Pearson's Linear Correlation Coefficient (PLCC), Root Mean Square Error (RMSE), and Mean Absolute Error (MAE). Higher SRCC and PLCC values and lower RMSE and MAE values indicate better results. We randomly divide the proposed \dataset dataset into 70\%, 10\%, and 20\% splits corresponding to training, validation, and test sets, respectively. We cross-validate 10 times to reduce bias and report the median test performance. 

\subsection{Results}

\begin{table}
    \centering
    \Large
    \resizebox{\linewidth}{!}{ 
    \begin{tabular}{llcccc}
    \toprule
    Method & Variant & SRCC & PLCC & RMSE & MAE \\ \midrule
    \multirow{2}{*}{DBCNN} & Base & 0.755 & 0.761 & 0.093 & 0.074 \\
     & \cellcolor{LightCyan70}\method & \cellcolor{LightCyan70}\textbf{0.776} & \cellcolor{LightCyan70}\textbf{0.780} & \cellcolor{LightCyan70}\textbf{0.090} & \cellcolor{LightCyan70}\textbf{0.069} \\
    \midrule
    \multirow{2}{*}{TOPIQ} & Base & 0.764 & 0.762 & 0.098 & 0.078 \\
     & \cellcolor{LightCyan70}\method & \cellcolor{LightCyan70}\textbf{0.808} & \cellcolor{LightCyan70}\textbf{0.805} & \cellcolor{LightCyan70}\textbf{0.088} & \cellcolor{LightCyan70}\textbf{0.069} \\
    \midrule
    \multirow{2}{*}{CONTRIQUE} & Base & 0.618 & 0.635 & 0.114 & 0.090 \\
     & \cellcolor{LightCyan70}\method & \cellcolor{LightCyan70}\textbf{0.631} & \cellcolor{LightCyan70}\textbf{0.651} & \cellcolor{LightCyan70}\textbf{0.106} & \cellcolor{LightCyan70}\textbf{0.083} \\
    \midrule
    \multirow{2}{*}{ARNIQA} & Base & 0.651 & 0.650 & 0.105 & 0.082 \\
     & \cellcolor{LightCyan70}\method & \cellcolor{LightCyan70}\textbf{0.687} & \cellcolor{LightCyan70}\textbf{0.672} & \cellcolor{LightCyan70}\textbf{0.103} & \cellcolor{LightCyan70}\textbf{0.079} \\
    \midrule
    \multirow{2}{*}{CLIP-IQA$^{+}$} & Base & 0.660 & 0.664 & 0.114 & 0.089 \\
     & \cellcolor{LightCyan70}\method & \cellcolor{LightCyan70}\textbf{0.666} & \cellcolor{LightCyan70}\textbf{0.666} & \cellcolor{LightCyan70}\textbf{0.107} & \cellcolor{LightCyan70}\textbf{0.084} \\
    \midrule
    \multirow{2}{*}{QualiCLIP} & Base & 0.665 & 0.642 & 0.124 & 0.097 \\
     & \cellcolor{LightCyan70}\method & \cellcolor{LightCyan70}\textbf{0.670} & \cellcolor{LightCyan70}\textbf{0.649} & \cellcolor{LightCyan70}\textbf{0.111} & \cellcolor{LightCyan70}\textbf{0.088} \\ \bottomrule
    \end{tabular}}
    \vspace{-5pt}
    \caption{Performance evaluation of NR-IQA methods on the \dataset dataset, trained using only the ground-truth labels (``Base") or also weak labels generated via the proposed \method (highlighted in cyan). Best performance for each method is highlighted in bold.}
    \vspace{-10pt}
    \label{tab:main_results}
\end{table}

We integrate the proposed approach with several state-of-the-art NR-IQA methods retrained for IISA. For a comprehensive analysis, we consider baselines based on diverse strategies. Specifically, we consider approaches that train a network from scratch with supervised learning (DBCNN \cite{zhang2018blind} and TOPIQ \cite{chen2024topiq}), methods that train a regressor head on top of an encoder pre-trained with self-supervised learning (CONTRIQUE \cite{madhusudana2022image} and ARNIQA \cite{agnolucci2024arniqa}), and VLM-based models that rely on prompt learning \cite{wang2023exploring} (CLIP-IQA$^{+}$ \cite{wang2023exploring} and QualiCLIP$^{+}$ \cite{agnolucci2024quality}). For each baseline, we train two variants on the \dataset dataset: a base with only ground-truth labels, and another that also uses weak labels from \method.

We report the results in \cref{tab:main_results}. We observe that our approach consistently improves the results, with relative performance gains up to 5\%. This proves that \method can be applied to any model, regardless of its specific characteristics, as a way to embed domain knowledge about the IISA task into the training process. Moreover, these results confirm that \cref{eq:iis_formula} can be leveraged in practice to enhance the value of every ground-truth IIS label  
by generating additional weakly labeled training data from it. Finally, we observe that, when combined with \method, TOPIQ achieves the best performance among the considered methods.

\subsection{Discussion}
\paragraph{Proposed method} We conduct controlled experiments to assess how key components affect our approach's performance, namely: 1) the number of weak labels $n_{wl}$; 2) the downscaling threshold $\rescalehyperparam$; 3) the interpolation algorithm used to generate the weak labels.
For each experiment, we keep other hyperparameters fixed unless otherwise stated. We consider only the TOPIQ model for its superior performance among the studied approaches.

\cref{tab:ablations} shows the results. Regarding the number of weak labels, the experiments show that increasing the value of $n_{wl}$ leads to better results up to $n_{wl}=2$, but beyond that performance declines. We hypothesize that too many weak labels introduce overly redundant information, as the image content remains constant across them, with only the scale varying. Likewise, a high downscaling threshold leads to the weak labels being overly similar to the ground-truth ones, thus reducing the performance improvement. Conversely, lower $\rescalehyperparam$ values allow the weakly labeled IIS to be higher (see \cref{eq:iis_formula}), possibly resulting in them being out of the ground-truth distribution (the average IIS in \dataset is 0.347). Finally, we observe that using the bilinear and bicubic interpolation algorithms to generate the weak labels leads to marginally lower performance than using the Lanczos one. This finding suggests that the interpolation algorithm has a limited impact in training models with \method. 

\begin{table}
    \centering
    \begin{tabular}{lcccc}
    \toprule
    Variant & SRCC & PLCC & RMSE & MAE \\ \midrule
    Base & 0.764 & 0.762 & 0.098 & 0.078 \\ \midrule
    $n_{wl} = 1$ & 0.803 & 0.801 & 0.090 & 0.072 \\
    $n_{wl} = 3$ & 0.788 & 0.785 & 0.096 & 0.077 \\ \midrule
    $\rescalehyperparam = 0.50$ & 0.795 & 0.780 & 0.097 & 0.076 \\
    $\rescalehyperparam = 0.80$ & 0.802 & 0.800 & 0.089 & \textbf{0.069} \\ \midrule
    Bilinear & 0.799 & 0.796 & 0.089 & 0.070 \\
    Bicubic & 0.803 & 0.803 & 0.093 & 0.074 \\ \midrule
    \rowcolor{LightCyan70} \method & \textbf{0.808} & \textbf{0.805} & \textbf{0.088} & \textbf{0.069} \\ \bottomrule
    \end{tabular}
    \vspace{-5pt}
    \caption{Performance evaluation of different variants of \method applied to the TOPIQ model \cite{chen2024topiq}. ``Base" represents the model trained only with ground-truth labels, without using our approach. Each subsequent variant involves altering only one hyperparameter at a time from the optimal configuration of \method (highlighted in cyan). Best performance is highlighted in bold.}
    \vspace{-5pt}
    \label{tab:ablations}
\end{table}

\paragraph{Assumption} 
In \cref{sec:iisa} we assume that the relationship between image quality and scale follows a concave-down or monotonic function. To verify this, we consider the KonX dataset \cite{wiedemann2023konx}, which provides quality annotations for images downscaled to three predefined resolutions. We find that 90\% of the image triplets (378 out of 420) show MOS trends across resolutions that are consistent with our assumption. We attribute the instances where the assumption does not hold to potential noise in the annotation process caused by subjective biases from annotators or interpolation artifacts. See the supplementary material for more details. Furthermore, studies that vary viewing distances while keeping image resolution fixed \cite{liu2014cid, gu2015quality, fang2015evaluation} also support our assumption. They show that image quality follows a concave-down function with respect to viewing distance.
In principle, changing the viewing distance at a fixed image resolution is equivalent to changing the resolution (and display size) at a fixed viewing distance. Together, these studies provide compelling evidence supporting the validity of our assumption across a wide range of conditions.

\section{Conclusions}
We defined the Image Intrinsic Scale (IIS) to explore the relationship between image quality and scale and introduced the IISA task. Our work puts forward a subjective annotation methodology for the IIS and creates the first dataset for IISA. The proposed approach leverages an IIS-specific property to generate weak labels. Experiments show that 
\method improves the performance of several NR-IQA models, suggesting its potential for enhancing future IISA approaches.
The IIS complements traditional quality ratings, providing a more comprehensive view of real-world image quality. Notably, the IIS introduces a unique advantage, as it offers a direct method to maximize image quality by downscaling. Since our contributions span all stages of development -- from task definition to predictive modeling -- our work lays a solid foundation for future research on the nuanced interplay between image scale and perceived quality.

\paragraph{Acknowledgements}
The dataset creation was partially funded by the Deutsche Forschungsgemeinschaft (DFG, German Research Foundation) – Project-ID 251654672 – TRR 161.

{
    \small
    \bibliographystyle{ieeenat_fullname}
    \bibliography{main}
}

\setcounter{section}{0}
\setcounter{subsection}{0}
\setcounter{figure}{0}
\setcounter{table}{0}
\setcounter{equation}{0}
\setcounter{footnote}{0}
\renewcommand{\thesection}{S\arabic{section}}
\renewcommand{\thefigure}{S\arabic{figure}}
\renewcommand{\thetable}{S\arabic{table}}
\renewcommand{\theequation}{S\arabic{equation}}

\maketitlesupplementary

\section{IISA Task: Additional Details}

\subsection{Sensitivity of the IIS}
\label{sec:sensitivity}

In this section, we compare the relative sensitivity of the quality ratings and the IIS. Here, by \textit{sensitivity} we refer to the precision of an annotation tool in detecting variations in image quality. The classical measurement tool in IQA is the rating scale, such as the 100-point ``continuous" scale or the discrete 5-point Absolute Category Ratings (ACR). In IISA we employ a 100-point scale corresponding to the rescaling values to measure the IIS. Here, we normalize both quality ratings and scale values to the interval $[0,1]$.

\begin{figure}[!t]
    \centering
    \includegraphics[width=\linewidth]{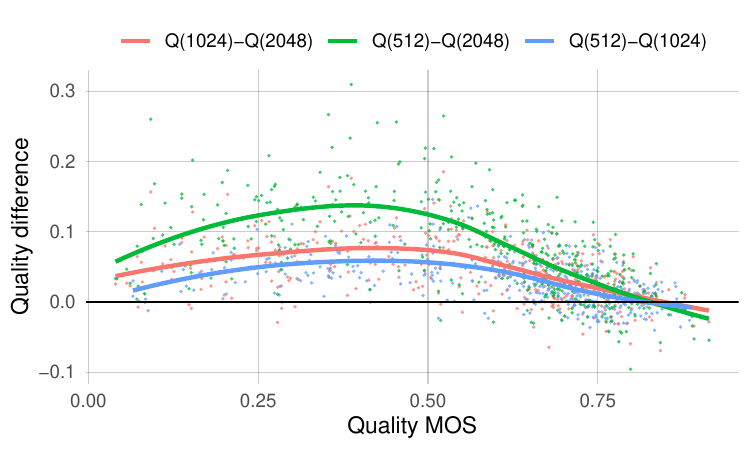}
    \vspace{-20pt}
    \caption{Quality differences between pairs of rescaled images belonging to the KonX dataset against the MOS of the higher-resolution image in each pair. The MOS for an image at a specific resolution is denoted as Q(image-width). We plot trend lines for each resolution pair.}    \label{fig:quality_vs_scale_konx}
    \vspace{-12pt}
\end{figure}

To conduct our analysis, we consider the KonX dataset \cite{wiedemann2023konx}. We aim to study the connection between the change in quality relative to the change in scale. Recall that KonX provides quality scores, in the form of a Mean Opinion Score (MOS), for the same images annotated at three resolutions: $512\times384$, $1024\times768$, and $2048\times1536$. We compute the quality differences between pairs of corresponding rescaled images across the three resolutions and plot them against the MOS of the higher-resolution image in each pair, as shown in \cref{fig:quality_vs_scale_konx}. 

We restrict our analysis to image scale pairs where the MOS at the higher resolution is below $0.85$. For these, the average quality increases with downscaling. The ratio of the resolutions in a pair gives the downscaling factor, which in turn gives the variation in scale, referred to as $\Delta S$. Thus, halving the resolution means $\Delta S = 0.5$ and on average corresponds to a quality increase (referred to as $\Delta Q$) of $0.038$, while downscaling to $S=0.25$ means $\Delta S=0.75$ and leads to $\Delta Q=0.076$. To measure the sensitivity, we employ the concept of leverage $\gamma$, defined as the ratio between the change in image scale and the change in image quality, \ie, $\gamma = |\Delta S|/|\Delta Q|$. Hence, $\gamma>1$ indicates that a large change in the scale results in a smaller change in quality, and vice-versa for $\gamma<1$. In our case, the leverage $\gamma$ is $6.6$ and $19.7$ for the two scale changes of $\Delta S=0.5$ and $\Delta S=0.75$, respectively. 

In Sec. \textcolor{iccvblue}{4.2}, we discuss the precision of subjective measurements, indicated by the confidence intervals of the aggregated annotations. The precision levels for IISA and NR-IQA are comparable. However, the leverage factor $\gamma>1$ implies that minor changes in quality result in larger variations in scale. Therefore, achieving a specific precision in measuring scale equates to an even finer precision in measuring quality. Thus, $\gamma$ acts as a leverage or magnifying factor. This indicates that the sensitivity of our annotation tool for IIS is higher when detecting subtle differences in quality compared to traditional quality rating scales, making IISA suitable for fine-grained quality assessment.

\subsection{IIS and Quality Scores}

We investigate the relationship between the IIS and quality scores, represented by MOS, by analyzing the overlapping images between the KonIQ-10k \cite{hosu2020koniq} and \dataset datasets.
KonIQ-10k comprises quality annotations for images downscaled to a fixed resolution of $1024 \times 768$ pixels. In contrast, the KonIQ-10k subset of \dataset contains the original high-resolution version of the same images, with the same content and aspect ratio but variable resolutions above $2048 \times 1536$ pixels.

\begin{figure}[!t]
    \centering
    \includegraphics[width=0.49\linewidth]{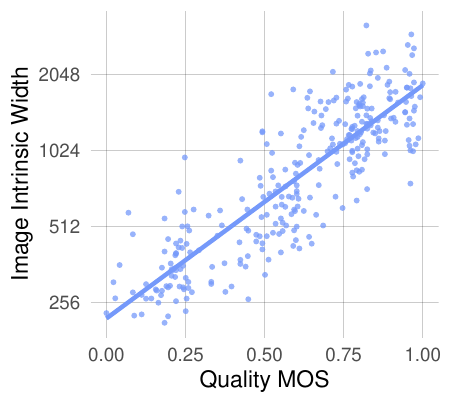}
    \vspace{-10pt}
    \caption{Relationship between image quality and intrinsic width (logarithmic scale) for the KonIQ-10k subset of \dataset.}
    \label{fig:intrinsic_width_vs_quality}
    \vspace{-5pt}
\end{figure}

By the definition of IIS, high-quality -- and thus presumably undegraded -- images should have an IIS of 1, as downscaling them can not reduce the visible degradation but merely results in a potential loss of details. Therefore, KonIQ-10k images with near-perfect quality (\ie, with the highest MOS) are expected to correspond to an IIS of 1 at a resolution of $1024 \times 768$ pixels. From another perspective, the original high-resolution versions of the images with the highest MOS should have an intrinsic width (\ie, the width of the image downscaled to its IIS) of \textit{at least} 1024 pixels. Indeed, such images could reach a near-perfect quality even when downscaled to a width larger than 1024 pixels. To validate this hypothesis, we plot the intrinsic widths of the images against their MOS in \cref{fig:intrinsic_width_vs_quality}. The results show that the images with the highest quality correspond to intrinsic widths higher than 1024 in almost all cases, thus confirming our hypothesis.

\begin{figure}[!t]
    \centering
    \includegraphics[width=0.49\linewidth]{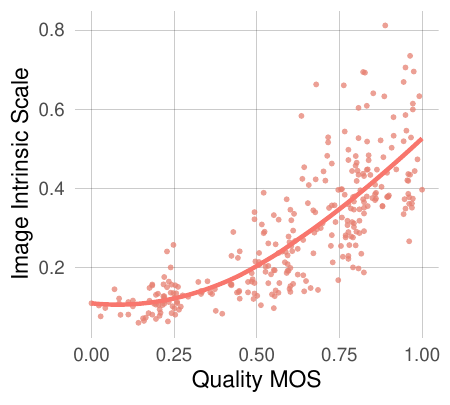}
    \includegraphics[width=0.49\linewidth]{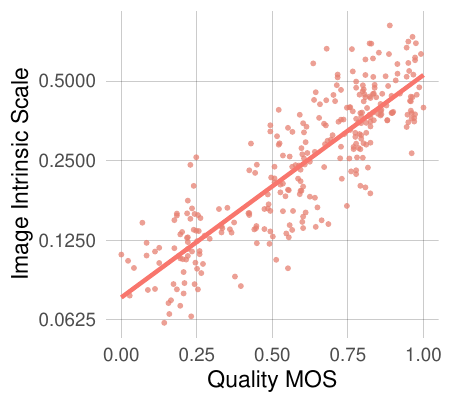}
    \vspace{-10pt}
    \caption{Relationship between image quality and intrinsic scale for the KonIQ-10k subset of \dataset, with IIS represented on a linear scale (left) and a logarithmic scale (right).}
    \label{fig:intrinsic_scale_vs_quality}
    \vspace{-5pt}
\end{figure}

In addition, we plot the IIS of the images against their MOS in \cref{fig:intrinsic_scale_vs_quality}. We observe a non-linear relationship between the IIS and quality MOS (\cref{fig:intrinsic_scale_vs_quality}, left). On the contrary, the logarithm of the IIS exhibits an approximately linear relationship with the quality scores (\cref{fig:intrinsic_scale_vs_quality}, right). This result emphasizes the different nature of the scales of the MOS and the IIS. Indeed, the quality ratings use a perceptually linear scale \cite{devellis2021scale}, while rescaling factors -- underlying the IIS -- are intrinsically non-linear.

\subsection{Discussion on Assumption}

In Sec. \textcolor{iccvblue}{6.3} we discuss our assumption that the relationship between image quality and scale follows either a concave-down or monotonic function. Specifically, prior works related to viewing distance \cite{liu2014cid, gu2015quality, fang2015evaluation} align with our assumption. 
In addition, we empirically test this assumption by analyzing the quality change with resolution in the KonX dataset \cite{wiedemann2023konx}. For each resolution, the estimated quality MOS have an average confidence interval of approximately 4.6\% relative to the rating range, demonstrating good precision and enabling us to draw the following conclusions.
First, we observe that for 90\% of the KonX images (378 out of 420) the MOS across the three resolutions supports our assumption. Second, for cases where the MOS follows a concave-up function across resolutions, due to the uncertainty of the MOS there is no single image for which we can assert with more than 90\% probability that our assumption does not hold. To determine this, we generate MOS values by resampling the individual quality ratings with replacement 100 times from the original pool of per-participant ratings. The fraction of samples that do not support our assumption provides the stated probability. We hypothesize that instances where our assumption appears not to hold may be due to subjective biases from annotators -- such as the presentation order of images during annotation, context effects like anchoring caused by the distribution of image quality within the same session, or individual interpretations of the quality scale -- as well as the presence of interpolation artifacts from downscaling, including aliasing, moiré patterns, or blurs.

\subsection{IISA Applications: Additional Details}

We present scenarios across industry and research where IISA is the perfect tool to optimize quality and resolution trade-offs.

\textbf{Printing and publishing} Printing an image too large can accentuate flaws present in the source image while printing too small sacrifices detail. IISA can guide the selection of print dimensions and resolution (dot-per-inch, DPI). This ensures consistently high-quality prints. Traditionally, this task is managed by an expert operator. IISA automates these decisions, allowing scalable deployment in online printing systems and enabling non-expert users to make optimal choices independently.

Moreover, web developers and UI designers often need to serve images across devices with different screen sizes and resolutions. Typically, responsive design uses fixed rules, whereas IISA enables content-awareness. For instance, an online image gallery can automatically size each photo based on its intrinsic scale. This ensures that users see images at the best quality for their device while saving bandwidth and load time.

\textbf{Gaming and graphics rendering} Modern games employ dynamic resolution scaling to maintain high FPS. However, choosing the amount of rescaling on each axis can dramatically affect quality. IISA offers a principled solution to this problem. Moreover, if the target FPS is not fixed, a game engine could automatically downscale rendered frames (slightly degraded by aliasing or motion blur) until just before quality starts dropping, ensuring players get the clearest visuals with optimal performance (FPS).

\textbf{Computational photography} Smartphone cameras rely on computational photography to balance resolution and noise. Although sensors may reach 100+MP, phones often merge pixels in low light to produce cleaner lower MP images -- effectively searching for the image’s intrinsic scale. IISA makes this process explicit and optimal. In tasks like super-resolution or denoising, algorithms can use IISA to determine when further resolution or noise reduction stops improving quality. 

\textbf{Benchmarking IQA methods} Traditional IQA metrics predict quality at a fixed resolution, while IISA requires accuracy at multiple scales. Our experiments show that off-the-shelf IQA models -- NR-IQA trained on traditional datasets -- perform poorly on the IISA task. By directly evaluating alignment with human perception across scales, IISA serves as a valuable benchmark. Performance improvements of NR-IQA methods on the IISA-DB benchmark indicate a deeper understanding of the quality–resolution trade-off, which is critical for both academic research and real-world image processing.

\textbf{Extending IQA study methodology} IISA introduces new methods for subjective image quality evaluation. Traditional IQA often has viewers rate an image’s quality at a fixed resolution, which can be difficult for subtle degradations in no-reference settings. By contrast, IISA asks viewers to resize an image until it ``looks best'', inherently comparing quality across scales. This approach improves sensitivity to minor artifacts (see Sec. \ref{sec:sensitivity}).

\textbf{Training dataset curation for image restoration} Constructing image datasets requires managing images of varying quality. Traditionally, low-quality images are discarded to avoid introducing erroneous priors into restoration models. However, IISA provides a more nuanced approach: rather than discarding low-quality images, it downscales them to their intrinsic scale to maximize quality. This preserves content diversity because discarding images purely on quality can disproportionately exclude dynamic or low-light scenes, which tend to be blurrier and noisier. By applying IISA, we mitigate that bias while maintaining a wider range of content.

\section{\dataset Dataset: Additional Details}
\subsection{Image Curation}

\begin{figure}[t]
    \centering
    \includegraphics[width=0.975\columnwidth]{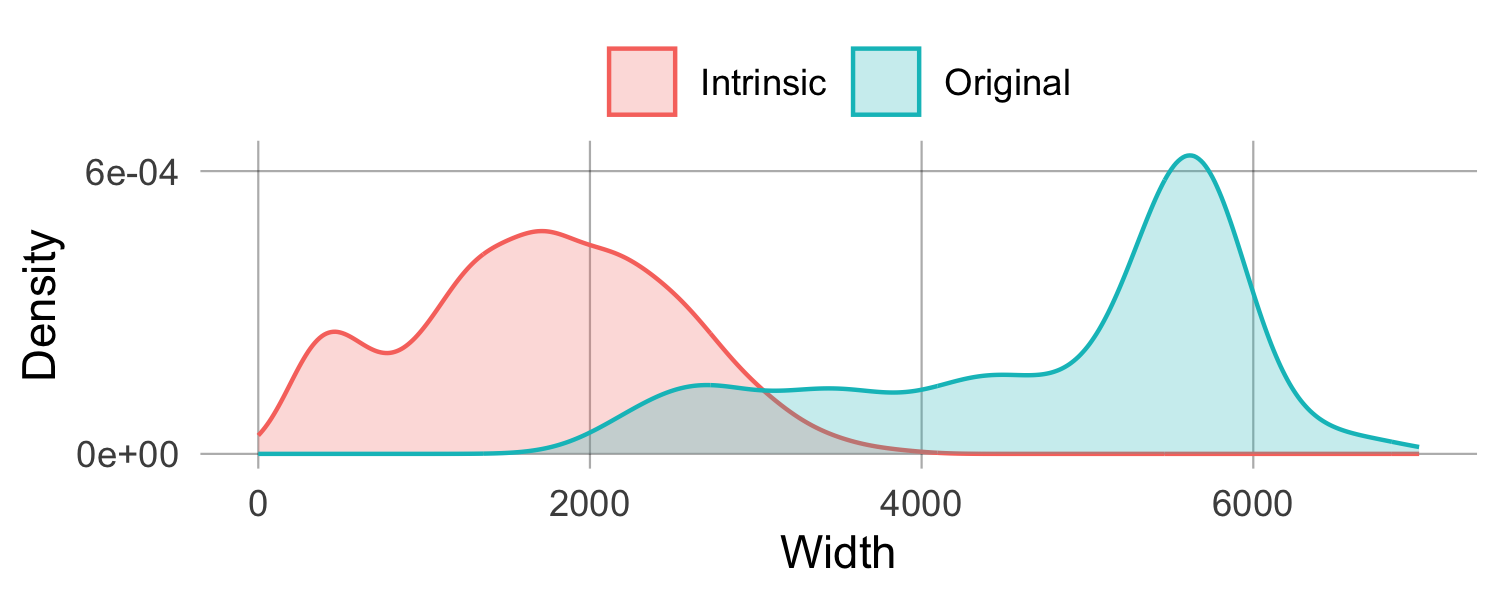}
    \vspace{-10pt}
    \caption{Distribution of the width of the images of the \dataset dataset at their original and intrinsic scale.} 
    \vspace{-10pt}
    \label{fig:dataset_density_width}
\end{figure}

To ensure the diversity of images in the proposed dataset, we selected images from two sources: 300 from the KonIQ-10k dataset \cite{hosu2020koniq} -- which were themselves sourced from Flickr -- and 600 from Pixabay. The two sets were chosen to balance the range of intrinsic resolutions (corresponding to the image rescaled to its IIS) in the database. The KonIQ-10k subset comprises lower-quality photos with smaller intrinsic resolutions, whereas the Pixabay images are typically of higher quality and intrinsic resolution.

Aiming for higher quality in the Pixabay subset, we selected newer camera models from a list of 71, focusing on those released after 2010 with full-frame sensors. From a pool of over one million images on Pixabay with EXIF information, we filtered for photos with a width greater than 4,000 pixels, resulting in approximately 18,000 images that met all criteria.  We sampled for diversity 600 images, maintaining uniform distributions among binned normalized favorites, likes, downloads, and user tags using a method similar to \cite{wiedemann2023konx}. Most of these were captured with \textit{Canon EOS 1/5/6D Mark 2/3/4} cameras, using various lenses and capture settings. For the KonIQ-10k subset, we also sampled for diversity regarding quality levels and machine tags -- with a confidence greater than 80\% -- using the same stratified procedure.

The last filtering step we applied was removing images containing identifiable people. Thus, we retained 248 images from KonIQ-10k and 537 from Pixabay. All images have a minimum width of $2048$ pixels and are annotated at their original resolution. \cref{fig:dataset_density_width} illustrates the distribution of image widths at their original and intrinsic scales. 

\subsection{Annotation Approach: Slider}
As explained in Sec. \textcolor{iccvblue}{4}, the subjective side of IISA is similar to the image Just Noticeable Difference (JND) task \cite{lin2022large}, which aims to determine the smallest level of distortion (\eg, compression amount) at which degradation becomes perceptible. This level is called the JND and is conceptually analogous to the IIS, which can be interpreted as the lowest downscale factor that maximizes the perceived quality of an image. Given the similarity between the two tasks, we took inspiration from the JND assessment to design the annotation strategy of the IIS. 

\begin{figure}[t]
    \centering
    \includegraphics[width=0.85\linewidth]{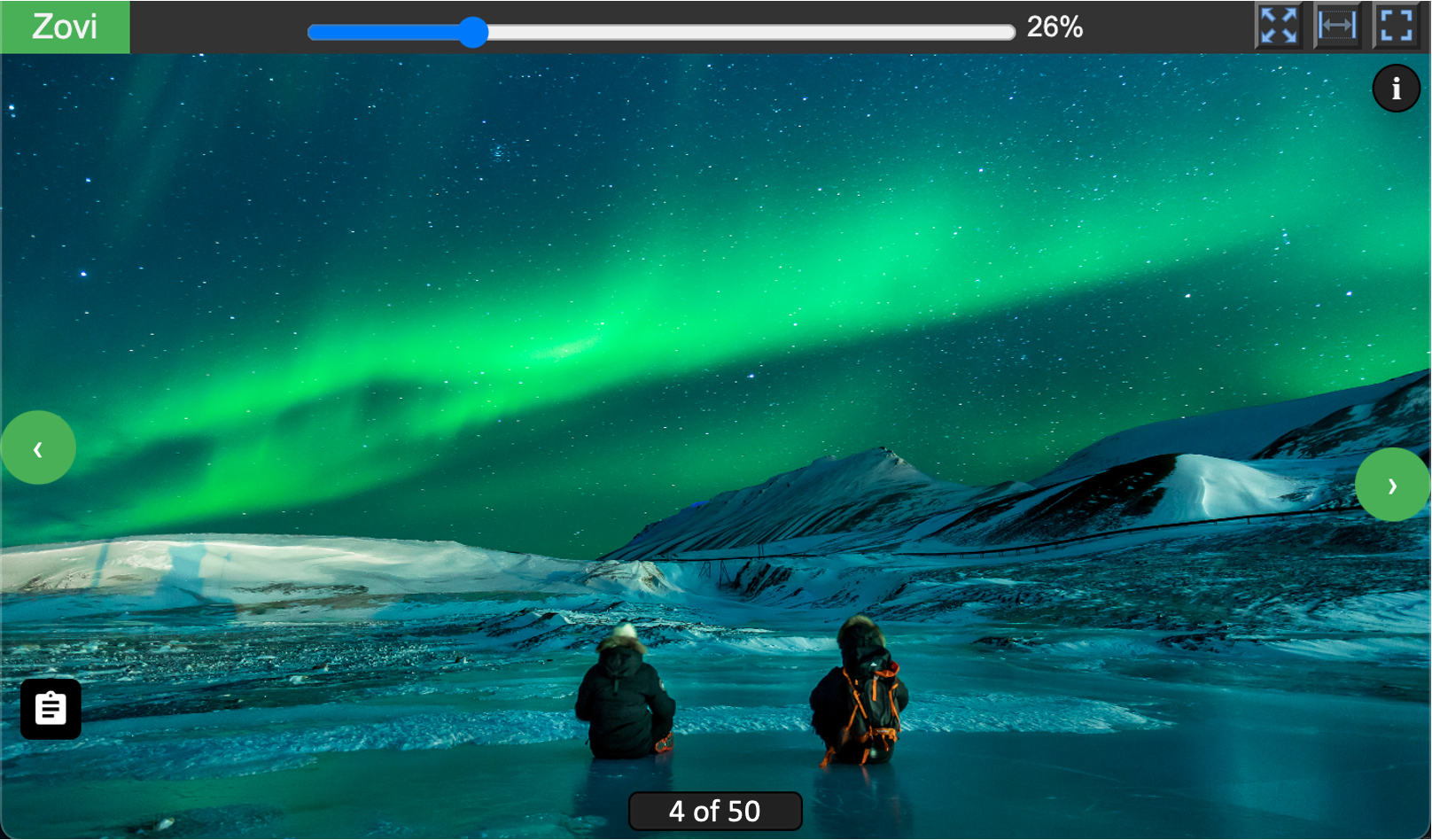}
    \caption{Screenshot of the UI of the ZOVI web application that we developed to annotate the IIS of an image.}
    \label{fig:zoom-viewer}
\end{figure}

The literature on JND assessment has proposed various annotation methods, including binary search and slider presentation \cite{lin2022large}. Among these, slider presentation has proven more effective, offering lower costs and higher precision. Therefore, as explained in Sec. \textcolor{iccvblue}{4}, we adopt a similar approach and develop an annotation tool (ZOVI) -- shown in \cref{fig:zoom-viewer} -- that displays a slider that allows the users to downscale the image from its original size ($scale=1$). In contrast, the binary search method is less efficient as it requires multiple independent participant judgments. For instance, if we were to consider 100 possible scale values the binary search would require $\lceil log_2{100} \rceil = 7$ comparisons. At a median of 3 seconds per judgment, it would take around 21 seconds to determine the IIS of an image. Empirical evidence suggests that the slider presentation is faster, taking 15 seconds per image, and provides more precise results for JND assessment \cite{lin2022large}.

\begin{figure}[!t]
    \centering

    \begin{minipage}{0.495\linewidth}
        \centering
        \includegraphics[width=0.485\linewidth]{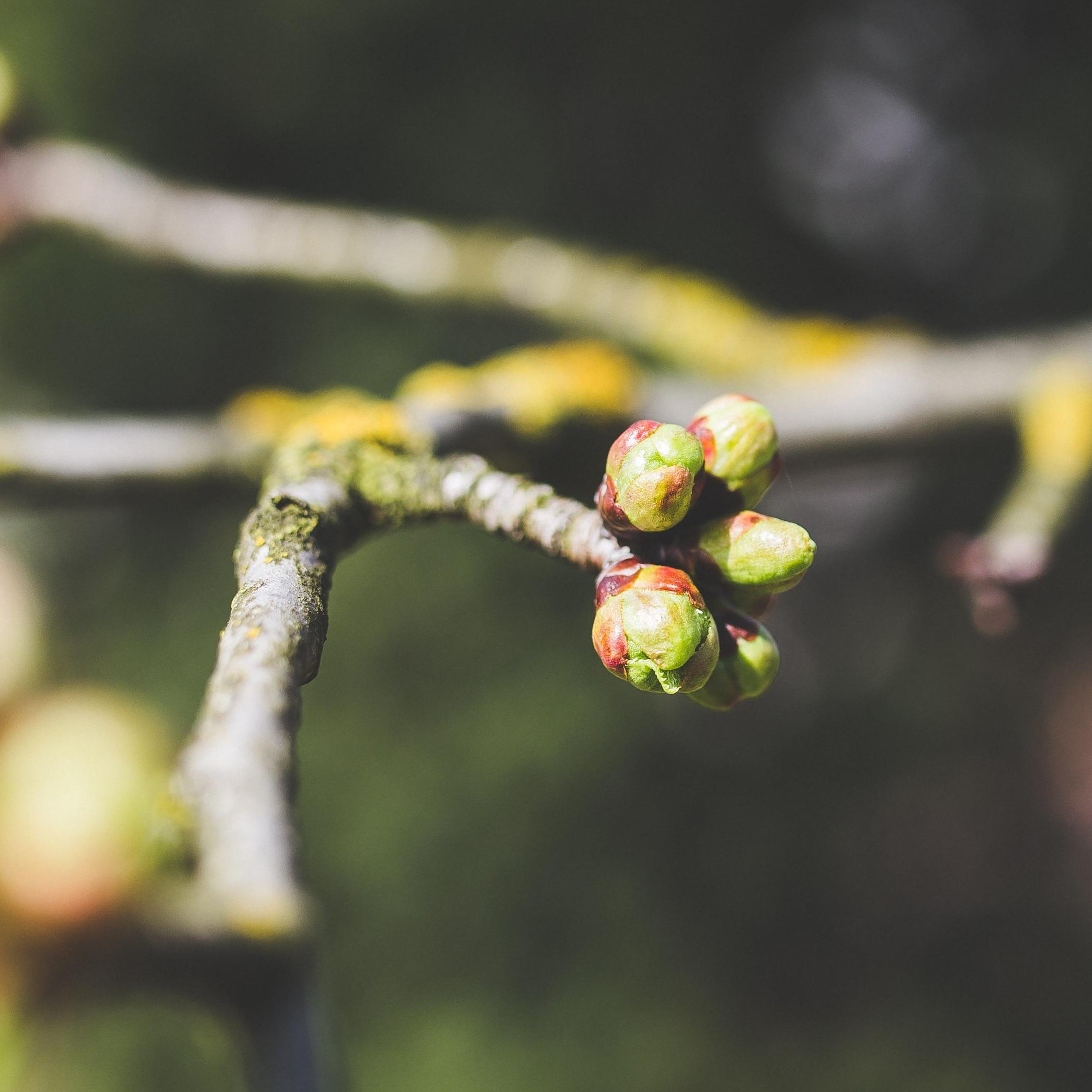}
        \includegraphics[width=0.485\linewidth]{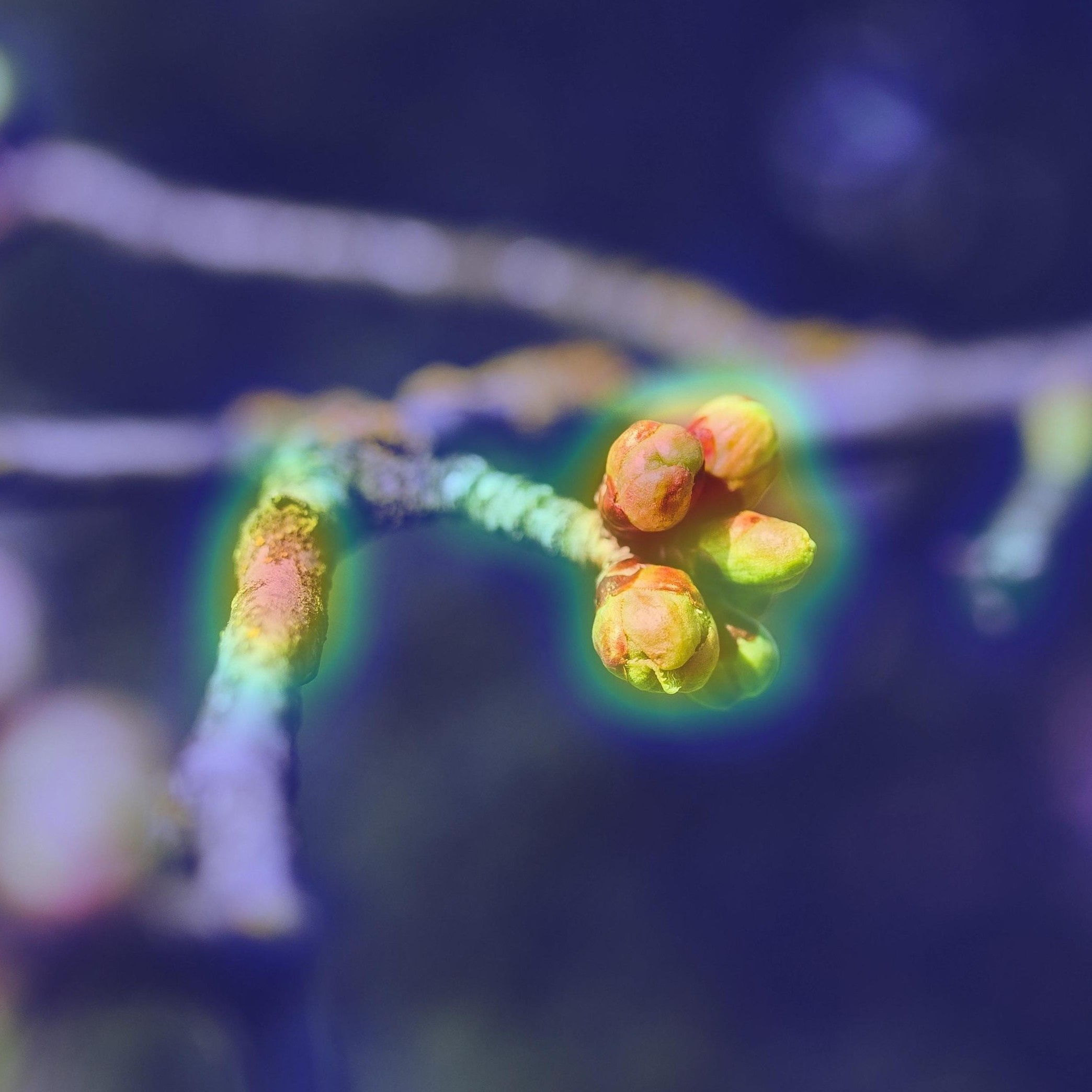}
        
        (a) 
    \end{minipage}
    \hfill
    \begin{minipage}{0.495\linewidth}
        \centering
        \includegraphics[width=0.485\linewidth]{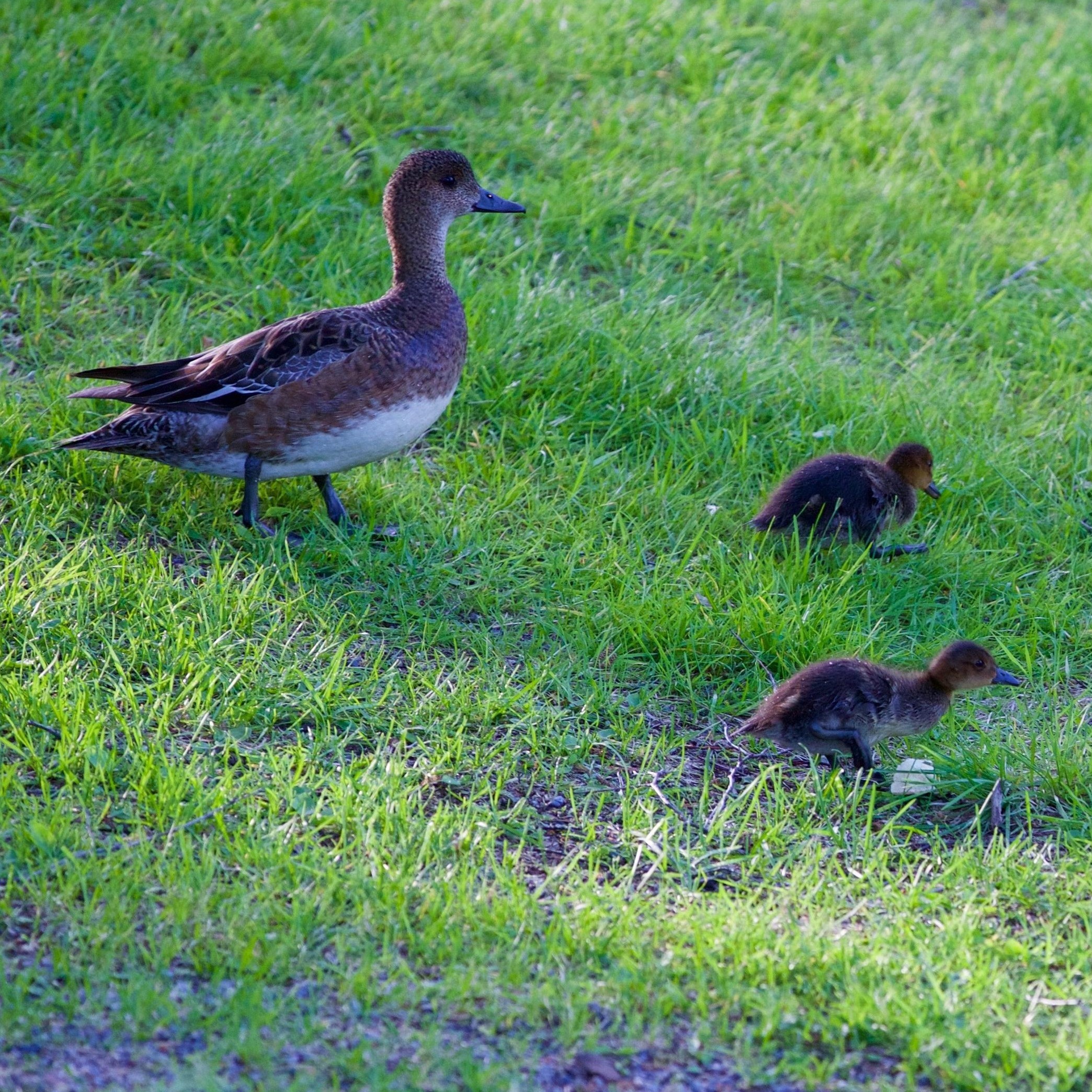}
        \includegraphics[width=0.485\linewidth]{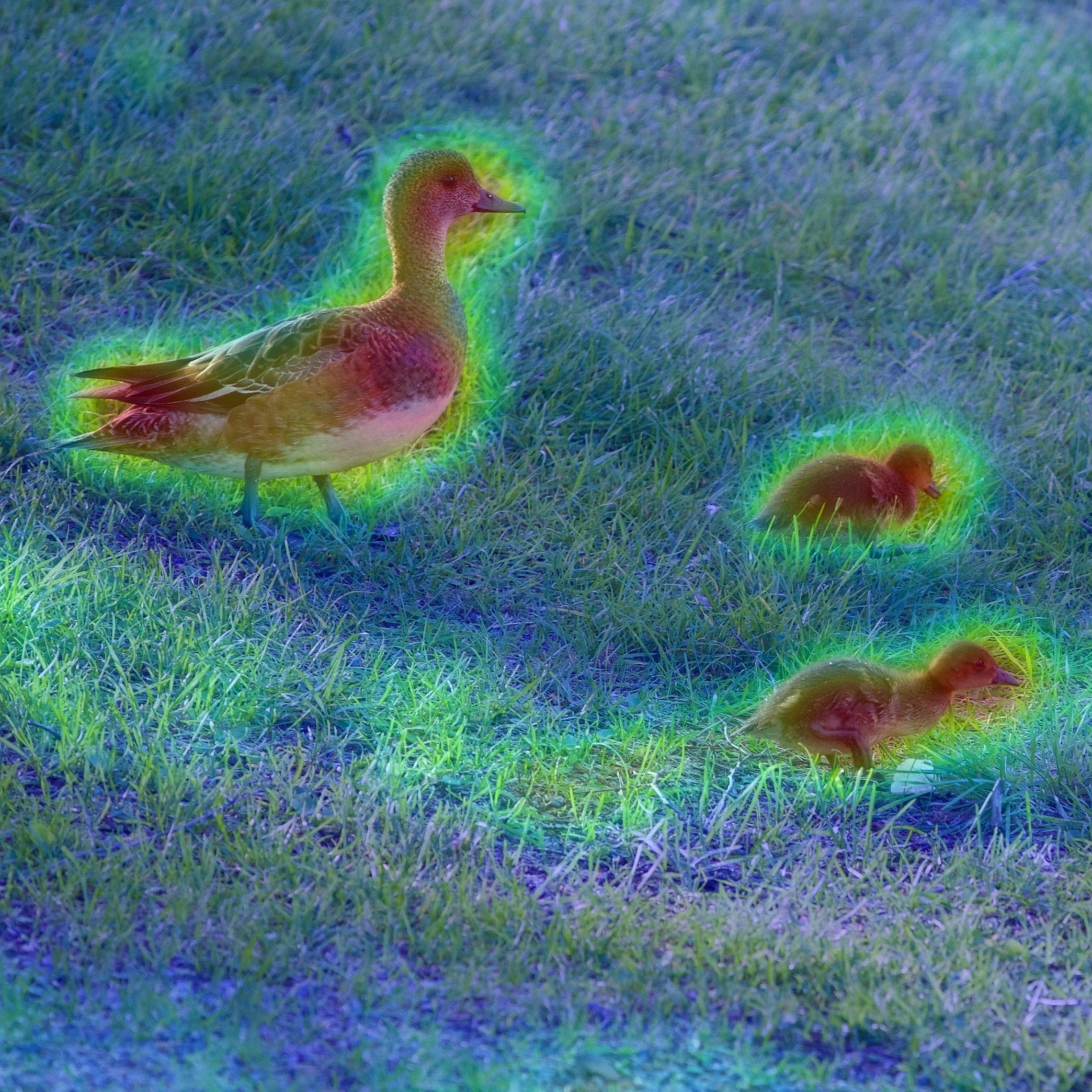}
    
        (b)
    \end{minipage}

    \caption{Visualization of TOPIQ's model attention.}
    \label{fig:topiq-attention}
\end{figure}

\subsection{Intrinsic Scale Aggregation Strategy}

Following the methodology detailed in Sec. \textcolor{iccvblue}{4}, we collect 20 IIS annotations for each image (10 participants $\times$ 2 opinions each). 
To obtain the ground-truth IIS labels we need a strategy to aggregate the single subjective opinions. We refer to this aggregated IIS value (\ie, the ground-truth one) as the Mean Opinion Intrinsic Scale (MOIS), drawing an analogy to the Mean Opinion Scores (MOS) used for assessing perceived image quality. One might na\"ively compute the arithmetic mean of the single IIS opinions, similar to how MOS are computed. However, the scale of the slider of our annotation tool is inherently non-linear. For instance, when an image's size doubles from 50\% to 100\%, the scale difference on the slider is twice that of when the image doubles from 25\% to 50\%. Therefore, values from different parts of the slider range should not be equally weighted, as plain averaging would. To address this, we apply a logarithmic transformation ($\log_2$) to the individual IIS values, which linearizes the scale before averaging. After averaging, we then exponentiate the result to revert to the original scale. This approach is equivalent to computing the geometric mean of the individual subjective opinions to obtain the MOIS of each image.

Formally, let $\Omega_j(I)$ be the $j$-th subjective opinion associated with the image $I$, with $j=1,\ldots,20$. Then, we compute the MOIS $\Omega(I)$ of the image $I$ by using the geometric mean:
\begin{equation}
\Omega(I) = 2^{\frac{\sum_{j=1}^N{log_2(\Omega_j(I))}}{N}} = \sqrt[N]{\prod_{j=1}^N{\Omega_j(I)}}
\label{eq:geom-mean}
\end{equation}
In this way, we account for the non-linearity of the slider scale. The final value $\Omega(I)$ represents the ground-truth IIS value of the image $I$, or MOIS. Across the 785 images composing our dataset, the average MOIS is 0.347, with per-image MOIS ranging from 0.060 to 0.811.

\subsection{Examples of Image-IIS Pairs}

\begin{figure*}[!t]
    \centering
    \includegraphics[width=\linewidth]{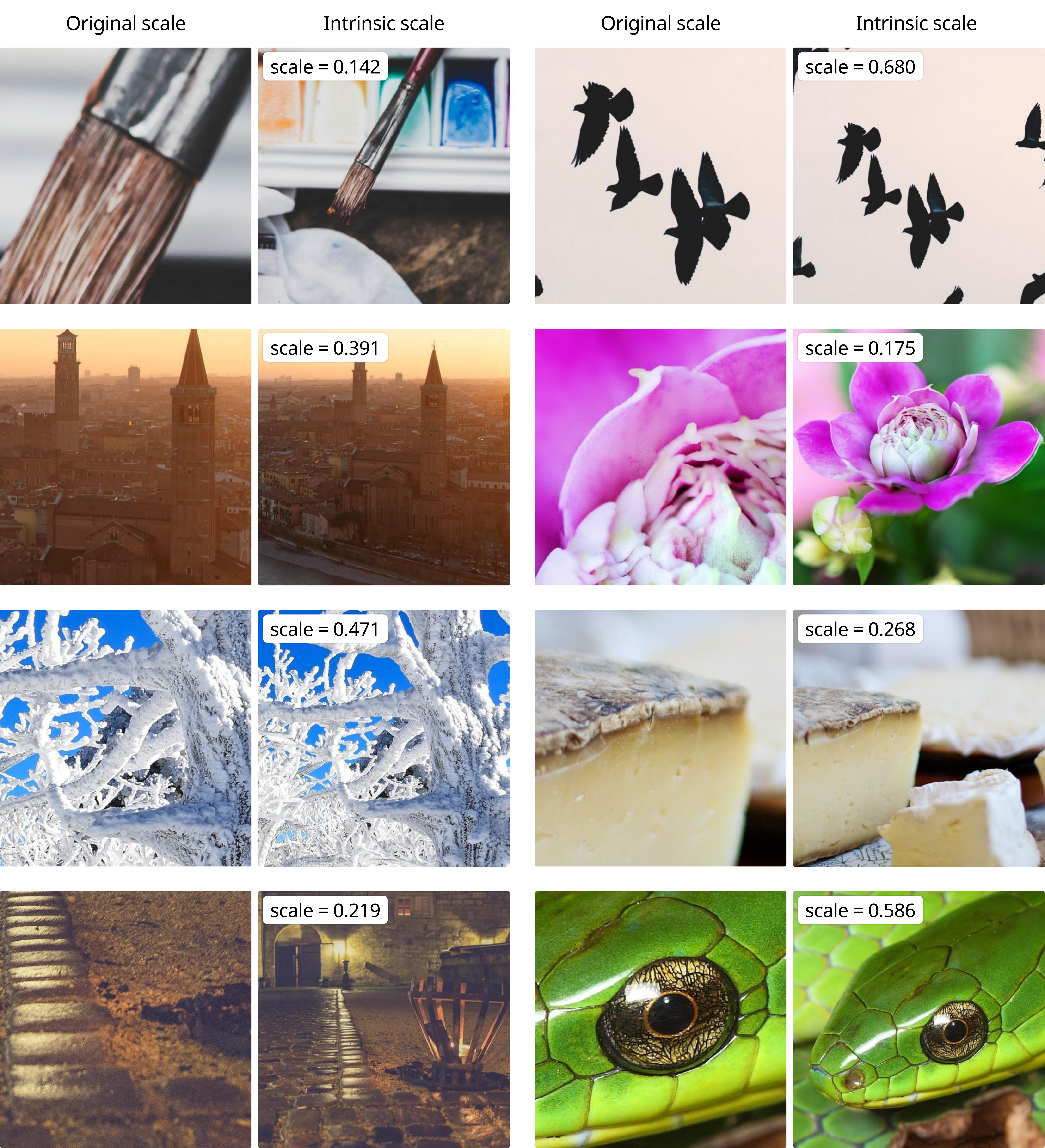}
    \caption{Examples of image-IIS pairs from the proposed \dataset dataset. Pairs of images are displayed in two columns. For each pair, the image on the left represents a crop from the original, while the image on the right depicts a crop from the original downscaled to the intrinsic scale. The content regions overlap between the two crops in each pair.}
    \label{fig:image-examples}
\end{figure*}

\dataset is designed to be diverse, featuring images with varying content and quality levels. \cref{fig:image-examples}  presents examples of image-IIS pairs from our dataset. Since we cannot display the images at their original size, we have cropped relevant sections and shown them at their original scale. Note that due to rescaling in the PDF viewer, the images may not appear exactly as they did to participants in the subjective study, \ie, at a 1:1 ratio of image to native screen pixels. However, the scale ratio between the original and intrinsic image crops remains consistent.

\subsection{Examples of Attention Maps}

We visualize the attention maps of the TOPIQ model trained with our WIISA approach in Fig. \ref{fig:topiq-attention}, using the method described in the original paper \cite{chen2024topiq}. Fig. \ref{fig:topiq-attention} (a) shows an image featuring a small foreground object in focus against a blurry background, while Fig. \ref{fig:topiq-attention} (b) depicts a high-semantic object surrounded by high-frequency texture. The model attends to in-focus and high-semantic regions, suggesting that IIS predictions are primarily driven by the degradation of high-level semantic content.

\section{Additional Experimental Results}

\subsection{Implementation Details}
During the training of each baseline, we extract square center crops with a size of 1536 pixels. We use data augmentation techniques that do not affect image quality, namely horizontal and vertical flips, with a probability of 0.5. We use Lanczos interpolation to generate the weakly labeled image-IIS pairs with our approach. We set the number of weak labels $n_{wl}$ to 2 and the downscaling threshold $\delta$ to 0.65. During testing, we feed each model the image at its original scale as input. We carry out the experiments on an NVIDIA H100 80GB GPU.

\subsection{Zero-Shot Multi-scale IISA}

Given the formulation of IIS reported in Eq. \textcolor{iccvblue}{1}, we can employ the quality scores predicted by a pre-trained NR-IQA method to estimate the IIS automatically. Specifically, given $n_s$ uniformly sampled scales $s$ in the range $[s_{lb}, 1]$, we can use a pre-trained NR-IQA model to assess the quality of each downscaled version $I^s$ of an input image $I$ and then find the scale for which the predicted quality is the highest. This would be an estimate for the IIS of image $I$.

Following the evaluation protocol described in Sec. \textcolor{iccvblue}{6.1}, we assess the performance of this zero-shot multi-scale approach on the proposed \dataset dataset. We employ two versions of the TOPIQ \cite{chen2024topiq} model pre-trained on the SPAQ \cite{fang2020perceptual} and UHD-IQA \cite{hosu2024uhd} datasets, which feature high-resolution images similar to those in \dataset. We use $n_s = 100$ scales and report the results in \cref{tab:supplementary_multiresolution_naive_results}. We observe that the zero-shot multi-scale approach achieves unsatisfactory performance, regardless of the pre-training dataset. We attribute this to NR-IQA models struggling to handle the change in perceptual quality caused by downscaling, as noted in \cite{wiedemann2023konx, huang2024high}. In addition, the multi-scale approach requires multiple model forward passes to obtain a single IIS prediction, which can be inefficient.

\begin{table}
    \centering
    \begin{tabular}{lcccc}
    \toprule
    Method & SRCC & PLCC & RMSE & MAE \\ \midrule
    TOPIQ (SPAQ) & 0.042 & 0.088 & 0.323 & 0.290 \\
    TOPIQ (UHD-IQA) & \textbf{0.054} & \textbf{0.166} & \textbf{0.290} & \textbf{0.255} \\
    \bottomrule
    \end{tabular}
    \caption{Evaluation of the performance on the \dataset dataset of the zero-shot multi-scale IISA approach based on the TOPIQ \cite{chen2024topiq} model. ($\cdot$) indicates the pre-training dataset. Best scores are highlighted in bold.}    \label{tab:supplementary_multiresolution_naive_results}
\end{table}

\section{Limitations}
The annotation process for IISA is time-consuming, requiring a median of 15 seconds per image versus 3 seconds for NR-IQA. Moreover, the significant effort and concentration required often make it challenging for typical crowdsourcing workers. In our pilot experiments, we found a high disqualification rate (about 90\%) among participants from Amazon Mechanical Turk, highlighting the need for more qualified but expensive expert annotators -- the latter participated in our experiments. Despite these challenges, the superior sensitivity of IISA justifies its use in scenarios requiring highly precise quality judgments. 

When collecting subjective IIS annotations, we employed Lanczos interpolation to rescale the images. While such an interpolation method guarantees high-quality results, different algorithms could be considered. For example, one could employ faster but lower-quality methods such as bilinear, or higher-quality, albeit slower, algorithms such as the full 2D Lanczos one.
While the experiments reported in Sec. \textcolor{iccvblue}{6.3} show that the interpolation algorithm does not make a significant difference for predictive IISA models, future work could extend our study by examining the effects of different interpolation methods for subjective IISA.

The scope of this study is focused on the impact of low-level distortions in User-Generated Content (UGC) images. Consequently, the applicability of our findings to specialized domains governed by different perceptual criteria, such as text-heavy images where readability is the primary quality indicator, remains largely unexplored. Extending the proposed framework to these areas is a promising direction for future work, which would involve adapting protocols to capture domain-specific quality factors.

\section{Future Work}
Our work suggests several promising directions for future research, including: 1) extending our dataset to incorporate more types of images, such as super-resolved, synthetically distorted, and computer-generated images; 2) conducting additional subjective studies -- similar to those of KonX -- to futher validate our assumption related to how image quality changes with scale; 3) analyzing the impact of different types of distortion on the IIS.

\end{document}